\documentclass[10pt, a4paper]{article}

\usepackage[final]{lrec2026} % this is the new style
\usepackage{url}
\definecolor{Red}{rgb}{0.89, 0, 0.06} 

\newcommand{\newtodo}[1]{\textcolor{Red}}

\definecolor{Puce}{rgb}{0.8, 0.53, 0.6} 

\newcommand{\newlf}[1]{\textcolor{Puce}}

\definecolor{Pistacchio}{rgb}{0.576, 0.773, 0.447}

\newcommand\blfootnote[1]{%
  \begingroup
  \renewcommand\thefootnote{}\footnote{#1}%
  \addtocounter{footnote}{-1}%
  \endgroup
}

\usepackage{longtable}
\defcitealias{allbus2023}{GESIS, 2025a}
\defcitealias{allbus2023compact}{GESIS, 2025b}

\title{German General Social Survey Personas: A Survey-Derived Persona Prompt Collection for Population-Aligned LLM Studies}
\name{Jens Rupprecht\textsuperscript{1,*}, Leon Fröhling\textsuperscript{2,*}, Claudia Wagner\textsuperscript{2,3,4}, Markus Strohmaier\textsuperscript{1,2,4}} 

\address{\textsuperscript{\rm 1}University of Mannheim, 
\textsuperscript{\rm 2}GESIS -- Leibniz Institute for the Social Sciences,\\
\textsuperscript{\rm 3}RWTH Aachen University, \textsuperscript{\rm 4}Complexity Science Hub}

\abstract{ % abstract needs to be between 150-200 words
The use of Large Language Models (LLMs) for simulating human perspectives via persona prompting is gaining traction in computational social science. However, well-curated, empirically grounded persona collections remain scarce, limiting the accuracy and representativeness of such simulations.
Here, we introduce the \emph{German General Social Survey Personas} (\emph{GGSS Personas}) collection, a \textit{comprehensive} and \textit{representative} persona prompt collection built from the German General Social Survey (ALLBUS).
The \emph{GGSS Personas} and their persona prompts are designed to be easily plugged into prompts for all types of LLMs and tasks, steering models to generate responses aligned with the underlying German population. 
We evaluate \emph{GGSS Personas} by prompting various LLMs to simulate survey response distributions across diverse topics, demonstrating that \emph{GGSS Personas}-guided LLMs outperform state-of-the-art classifiers, particularly under data scarcity. Furthermore, we analyze how the representativity and attribute selection within persona prompts affect alignment with population responses. Our findings suggest that \emph{GGSS Personas} provide a potentially valuable resource for research on LLM-based social simulations that enables more systematic explorations of population-aligned persona prompting in NLP and social science research.
\\ \newline  
}

\begin{document}

\maketitleabstract

\section{Introduction}
\label{sec:introduction}
\vspace*{-0.6\baselineskip}
\blfootnote{*~Equal contribution.}
Simulating human behavior represents a complex challenge given the vast diversity of human experiences and perspectives. Traditional research methods, such as large-scale question-oriented surveys, are crucial for informing social science research and policy decisions, but they are often costly and time-consuming to administer. As a result, researchers are increasingly turning to innovative approaches to capture the nuances of human behavior. One recent avenue is the use of large language models (LLMs) and so-called persona prompting \cite{chen2024from, tseng2024two, lutz2025prompt, zhang2025personalization}, i.e., the use of personas descriptions to induce role-playing in LLMs, as an accessible option to steer and control simulated human behavior in surveys \cite{ma2025algorithmic, miranda2025simulating}.

\textbf{Persona prompting} refers to a broad range of prompting techniques that aim to harness the general world knowledge captured by LLMs and steer the generation of simulated survey responses toward the perspective of the persona introduced in the prompt. Compared to alternative approaches for the alignment of LLMs, such as training dedicated models \cite{feng2023pretraining} or fine-tuning existing ones \cite{orlikowski2025beyond, suh2025language}, prompt-based approaches offer a number of crucial advantages: they are \textit{modular} (i.e., they can be used with different models and for different tasks), \textit{upgradable} (i.e., they can be modified and can be expected to improve in performance with the release of improved models), and \textit{shareable} (i.e., the prompts can easily be made available to other potential uses and users), among other advantages. 
%However, recent advances in persona prompting have been held back by the lack of well-curated and validated persona collections, which are essential for generating accurate and representative simulations.

\begin{figure*}[ht]
    \centering
    \includegraphics[width=\linewidth]{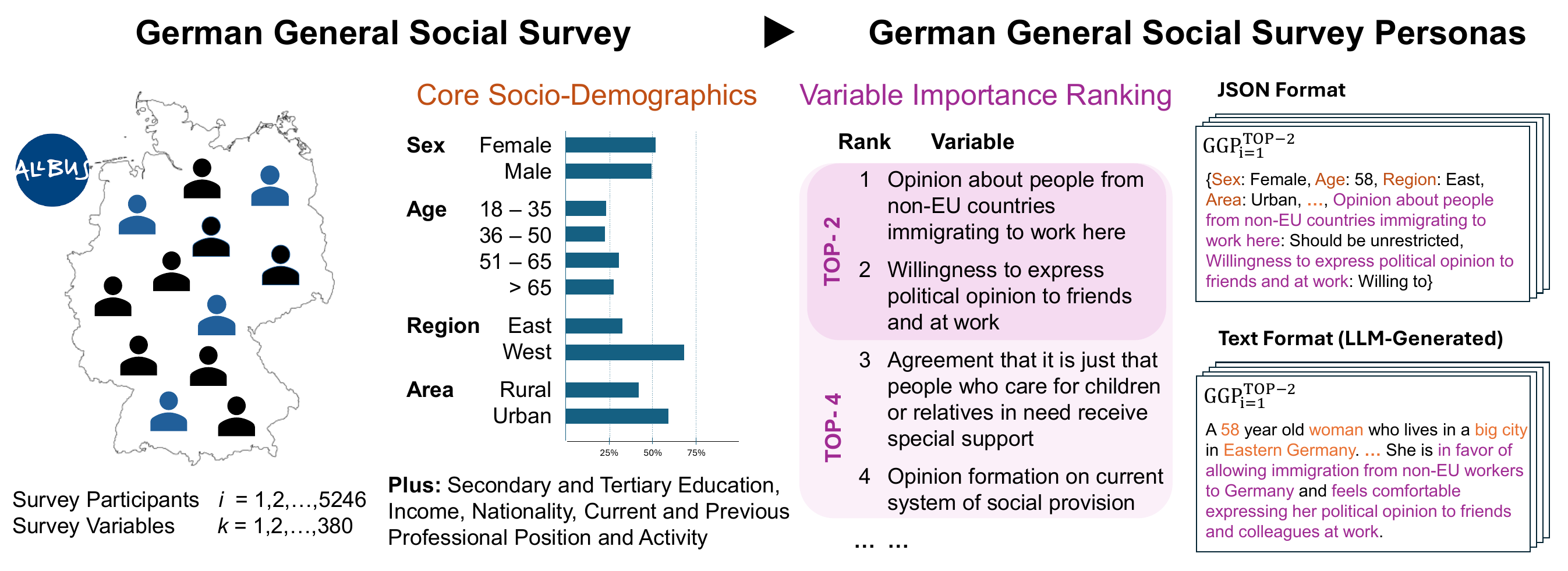}
    \caption{
    \textbf{Grounding the \emph{German General Social Survey Personas} (\emph{GGSS Personas}) in the ALLBUS survey.} We construct individual persona prompts for each ALLBUS participant, varying size and composition via available attributes. A global variable importance ranking informs the selection of the $k$ most important attributes (TOP-$k$). Personas comprise a fixed block of core socio-demographics and a more extensible block of TOP-$k$ attributes that allow varying information content. The \emph{GGSS Personas} is available in JSON- and full-text formats. The originally German survey items and personas are presented in English for illustration.}
    \label{fig:figure_1}
\end{figure*}

\textbf{\emph{German General Social Survey Personas}} are a prompt collection that consists of $5,246$ personas constructed from a representative sample of the German population. By grounding the \emph{German General Social Survey Personas} collection in the information available in the ALLBUS \citepalias{allbus2023}\textemdash the German General Social Survey and thus the premier source of information on the attributes and attitudes of the German population\textemdash we demonstrate that more systematic and principled approaches to the construction of persona prompt collections are possible~\footnote{German General Social Survey Personas are available after registration: \url{https://doi.org/10.4232/1.14707}}. 

\emph{GGSS Personas} offers various potential advantages: First, personas introduce relevant contextual information to the model, providing it with the necessary information to meaningfully anchor predictions for a task or target variable in empirically observed associations and connections. 
% Where previous work has limited persona prompts to the use of a small set of core socio-demographic attributes \cite{orlikowski2025beyond}, rarely extended with few variables selected for a specific task and context (e.g., politics) \cite{von2025vox, holtdirk2025learning}, or with psychological personality characteristics \cite{castricato2025persona}, 
%We construct \emph{German General Social Survey Personas} through a data-driven method that identifies the most important variables from the full range of information covered in the ALLBUS. 
Second, the ALLBUS is a probability-based survey and the personas derived from it aim to represent the German population. While there has been a focus on the biased representation of populations that LLMs default towards in their generation of survey responses \cite{santurkar2023whose, durmus2024towards, sen2025missing}, the \emph{GGSS Personas} may potentially help to align LLMs better with the German population. 
%enables exploration of how representative persona collections could help to achieve a more balanced representation of the population.

As a main contribution, this paper introduces \emph{German General Social Survey Personas}\textemdash a novel textual resource for NLP and Computational Social Science (CSS). We describe the creation process of \emph{GGSS Personas} and offer a preliminary analysis of its potential and utility through two exemplary demonstrations.
%First, we investigate how the number and choice of survey attributes reflected in the persona descriptions impact the ability of LLMs to simulate response distributions of the German population for randomly selected survey variables. Second, we compare the use of the representative \emph{GGSS Personas} to artificially created persona collections that oversample on selected characteristics such as income or ideological leaning as well as widely used persona collections from the literature such as the \textit{PersonaHub} \cite{ge2024scaling} to explore the role of representativity on the alignment of the generated response distributions with the population response distribution.

% \textbf{Structure of this Paper:} We provide an overview of the use of persona collections in different CSS and NLP tasks in Section~\ref{sec:related}. In Section~\ref{sec:GGSS Personas}, we describe the pipeline we used for creating the \emph{GGSS Personas} from the German General Social Survey. In Section~\ref{sec:empirical}, we first describe our preliminary analysis to demonstrate the use of the \emph{GGSS Personas} before presenting the results. We end with a discussion of representative persona collections as a novel type of resource (Section~\ref{sec:discussion}) as well as their limitations (Section~\ref{sec:limitations}).

\section{Related Work}
\label{sec:related}

Persona prompting is a \textbf{versatile and accessible method for social simulations} and pluralistic alignment with LLMs \cite{anthis2025position, sorensen2024position}. There is already a wide range of applications and promising demonstrations across disciplines and tasks, including (but not limited to): vote choice prediction \cite{argyle2023out, von2025vox}, generation of synthetic public opinions \cite{hwang2023aligning,ma2025algorithmic, miranda2025simulating}, annotation of (subjective) constructs \cite{hu2024quantifying, beck2024sensitivity, frohling2025personas}, simulations of participants in economic and social science experiments \cite{aher2023using, hewitt2024predicting}, including for underrepresented and otherwise hard to reach populations \cite{gonzalez2025llms}, simulation of pluralistic debates \cite{ashkinaze2025plurals}, behavioral simulation of users \cite{chen2025personatwin, salem2025tinytroupe}, and even productive real world tasks such as red teaming \cite{deng2025personateaming}. Systematic reviews of the literature on the use of LLMs for role-playing and persona prompting provide an overview of this dynamic area of research \cite{lutz2025prompt, zhang2025personalization, chen2024from, tseng2024two}.

\textbf{Beyond purely prompt-based approaches}, \citet{kim2023ai} propose a framework that fine-tunes LLMs on survey data for predicting responses at an individual level, and \citet{cao2025specializing} and \citet{suh2025language} use group-level information to fine-tune LLMs for the simulation of response distributions. \citet{orlikowski2025beyond} compare the use of personas for (zero-shot) persona prompting with fine-tuning LLMs on pairs of annotators' persona descriptions and their annotations, finding that fine-tuning strictly outperforms zero-shot in simulating individual-level annotations. Apart from fine-tuning, \citet{chen2025persona} introduce the idea of persona vectors, extracted via persona descriptions and used to monitor and steer LLM generations. \citet{sun2025persona} propose an approach that leverages personas via retrieval-augmentation to personalize LLM responses.

\textbf{Conceptual criticism} of the use of persona prompting include the work by \citet{cheng2023marked,cheng2023compost}, who show that simulations of political and marginalized groups tend towards caricature, and that LLM-generated personas of non-white, non-male demographics exhibit patterns of othering and exoticization. Similarly, \citet{wang2025large} show how LLMs are likely to misportray and flatten the representation of demographic groups. \citet{kim2024persona} investigate how the use of personas might degrade performance by distracting the model from the task at hand. \citet{kirk2024benefits} discuss the broader context and societal implications of aligning LLMs with individuals.

% Existing persona collections that have been used extensively, e.g., the PersonaHub Ge et al. (2024), are mostly convenience collections that have been created without empirical grounding and regard for methodological rigor or simulation precision \citet{li2025...}. 

In the literature, there are some noteworthy demonstrations of how \textbf{well-curated, empirically-grounded persona collections} lead to improved performance in their applications.
\citet{park2024generative} show how the use of transcripts of hour-long interviews with participants, i.e., extremely information-rich persona context, leads to impressive LLM performance in predicting individuals' survey responses. Similarly, \citet{toubia2025database} survey a representative sample of US-based crowdworkers to collect extensive information on their demographic-, psychological-, economic-, personality-, and cognitive measures, and \citet{peng2025mega} show that the use of this detailed personal information improves the correlations between human and LLM-simulated responses. 

However, this quality comes at the price of up to \$$100$ paid to $1,052$ participants for hour-long interviews and \$$37$ paid to $2,058$ crowdworkers for answering $500$ survey questions, respectively. Using \textbf{freely and publicly available survey data} offers an accessible alternative for grounding persona collections in high-quality empirical information. Most similar to our work, \citet{castricato2025persona} introduce PERSONA, a collection of $1,586$ synthetic personas created from US census data. In contrast to their approach, our \emph{GGSS Personas} establishes 1:1 correspondence between survey respondents and persona prompts, and automatically identifies important persona attributes from the full range of survey variables instead of limiting selection to a set of $31$ variables.

With the introduction of \emph{GGSS Personas}, we aim to contribute a well-curated, empirically-grounded persona collection as a crucial resource for meaningful advances in research on persona-based LLM personalization.

\section{\emph{German General Social Survey Personas}}
\label{sec:ggp}

\emph{German General Social Survey Personas} is derived from the ALLBUS, the German General Social Survey (Figure~\ref{fig:figure_1}). While the \emph{GGSS Personas} is necessarily tied to the German population at a specific moment in time, we propose a creation process that can be applied to different (population) surveys in order to create survey-derived persona collections for different contexts and purposes.

\paragraph{Grounding \emph{German General Social Survey Personas}}

%The \emph{German General Social Survey Personas} is empirically grounded in the German General Social Survey, the ALLBUS. 
Ideally, a population survey panel used for the construction of persona prompt collections should cover a set of core demographics and additional attributes for a representative sample drawn from the population of interest. Following the discussion on the use of the terms \textit{representative sample} and \textit{representativity} in the scientific literature by \citet{kruskal1979representative}, we believe that population-level general social surveys such as the ALLBUS in Germany, the General Social Survey (GSS) in the US, the British Social Attitudes Survey (BSA) in the UK, or the European Social Survey (ESS) covering 31 European Countries in its latest round, could represent plausible empirical foundations for representative persona prompt collections.

To create this first version
% ~\footnote{We clearly identify our persona collection by referencing the source and year: \emph{GGSS Personas-2023}.} 
of the \emph{GGSS Personas}, we use the latest release of the biannual ALLBUS \citepalias{allbus2023}, released in January 2025 and including responses collected between April and September 2023. It features $5,246$ participants and a total of $605$ variables. We work with the ALLBUScompact~\footnote{\url{https://search.gesis.org/research_data/ZA8832}} \citepalias{allbus2023compact}, the openly available version of the ALLBUS, featuring $579$ variables. For data protection reasons, the ALLBUScompact contains information in reduced detail on the socio-demographic characteristics age, origin, occupation and income, and omits all fine-grained geographic variables.

% While the exact meaning of \textit{representativity} is contested and the corresponding claim delicate in survey research \cite{cornesse2018there}, we base our conceptualization on the discussion offered by \citet{kruskal1979representative}. They observe that in the scientific literature the vague use of the concept may be warranted, even profitable, as long as it is backed up and specified by some method of probability sampling. 

% The latest available technical report for the ALLBUS \cite{wasmer2017konzeption} details the probability sampling used to draw the sample of ALLBUS participants from the general population. Based on population register data, a two-step disproportional random sampling method is used. The first step of the sampling procedure randomly samples a proportional number of sample points across all German municipalities. A second step then randomly draws individuals registered in the sample point municipalities. While the sampling procedure is fully probabilistic, it actively oversamples individuals living in East Germany to ensure sufficient sample sizes for fine-grained analyses. 

In our preliminary analysis, we explore the effects of using a representative persona collection such as the \emph{GGSS Personas}\textemdash representative of the German population by virtue of mirroring all ALLBUS participants\textemdash as well as unrepresentative variants of it, created by systematically oversampling on selected attributes (Section~\ref{sec:empirical}).

\begin{table}[t]
    \centering
    \footnotesize
    \begin{tabular}{ccccc}
        \hline
        \textbf{TOP-$k$} & \textbf{Average Attributes} & \textbf{Full Personas} \\
        \hline
        2 & 1.34 & 3374 \\
        4 & 2.68 & 1848 \\
        8 & 5.09 & 821 \\
        16 & 10.52 & 0 \\
        32  & 21.28 & 0 \\
        64 & 38.80 & 0 \\
        128 & 77.75 & 0 \\
        256 & 157.54 & 0 \\
        380 & 242.27 & 0 \\
        \hline
    \end{tabular}
    \caption{\textbf{Completeness of persona descriptions.} While the TOP-$k$ value in each row represents the maximum possible number of attributes featured in the collection's personas, the actually included average number is always lower due to some variables being unavailable for each survey respondent.}
    \label{tab:persona_attributes}
\end{table}

\paragraph{Attribute Selection}

Besides the choice for the population survey that serves as the fundament of a representative persona prompt collection, the second relevant selection concerns the set of survey variables that are included in the persona descriptions. These persona attributes should be chosen in a way that they provide the (most) relevant context to the LLM for it to make an informed prediction of the individual's attitude or behavior for a given task. 

While it is theoretically feasible to use all $579$ variables featured in the latest release of the ALLBUScompact in the persona prompts, this would unavoidably come at the price of increased context window requirements, computational resources, as well as potentially degrading performance due to the dilution of relevant information or the failure of the model to put equal attention to all variables in a lengthy persona description. For example, \citet{shi2023large} have shown that LLMs can easily be distracted by irrelevant information included in the context window, and \citet{de2025principled} have observed performance drops after including irrelevant persona details in persona prompts.  

% Existing work primarily includes sociodemographic \cite{orlikowski2025beyond} and small numbers of additional, mostly political attributes in their personas \cite{von2025vox,cheng2023compost,ashkinaze2025plurals,argyle2023out}, or includes complementary information such as psychological inventories \cite{castricato2025persona}. Other persona collections are created without any empirical grounding \cite{ge2024scaling}.

One of the core advantages of \emph{general} population surveys such as the ALLBUS is the coverage of a variety of different topics, ranging from core demographic variables such as age and gender to very specific, thematic variables covering, e.g., the participant's political views or lifestyle choices. Therefore, we design a data-driven method that allows us to identify and select the statistically most important variables, instead of having to (artificially) restrict the \emph{GGSS Personas} to cover only task-specific areas such as socio-demographics, psychological inventories, or political beliefs.

From the $579$ available survey variables, we exclude those that are purely technical interview paradata (e.g., information on the interviewer), those that we identify as core socio-demographic variables and include in all persona descriptions (see Figure~\ref{fig:figure_1}), and those that we randomly select as outcome variables (see Section~\ref{sec:empirical}). For each of the $406$ remaining variables, we fit a Random Forest Classifier, using all other $405$ variables as predictors. For each fitted model, we identify the ten most important predictors using their feature importance scores, and aggregate those scores for each predictor to create a variable importance ranking. This ranking features $380$ variables after dropping all variables that never appeared among the top ten most important features for any fitted model.

We expect that variables with higher aggregate feature importance score explain more variance of the other variables in the survey. This ranking thus allows us to create versions of the \emph{GGSS Personas} collection for different computational budgets and needs by selecting only the $k$ most important attributes (\textit{TOP-$k$}). In situations with restricted prompt length or inference time, versions with less attributes may be preferable, while in unrestricted settings, the number of included attributes may be increased, up to the inclusion of all $380$ survey variables. In our preliminary analysis, we explore the effects of increasing the number of persona attributes on the ability of LLMs to accurately simulate survey response distributions (Section~\ref{sec:empirical}).

An important limitation of our approach is its inability to deal with missing values. Because large surveys such as the ALLBUS work with questionnaire splits to manage the number of questions each participant is asked to answer, no participant will have answered all survey questions. Thus, some information is just not available for some of the participants, meaning that for any given variable some persona descriptions will always be incomplete, missing individual attributes. Table~\ref{tab:persona_attributes} shows the extent to which the persona descriptions across the different TOP-$k$ collections are complete. The $k$ thus represents the maximum number of attributes possibly featured in a persona of a given TOP-$k$ collection, and the reported average is the actual number of attributes included. We discuss the implications for the \emph{GGSS Personas} and our preliminary analysis in Section~\ref{sec:limitations}.

\paragraph{Persona Generation}

In the final step of our pipeline, we turn the persona attributes into a format that can easily be plugged into any LLM prompt to explicitly set the persona perspective that the model is supposed to take on during response generation.

There is a broad range of different prompting techniques in the literature \cite{lutz2025prompt}, ranging from the use of key-value pairs in a dictionary or \textbf{JSON-format} (e.g., \citet{castricato2025persona}) to the use of \textbf{full-text} persona descriptions (e.g., \citet{ge2024scaling}), oftentimes generated using LLMs. While personas in the JSON-format are easier to construct and more convenient to work with, the full-text persona descriptions resemble the natural language interactions that especially instruction-tuned LLMs have been optimized for more closely.

We evaluate the impact of using different persona formats for predicting survey response distributions. Since we do not observe a large impact of the persona format on the experimental results, (see Appendix~\ref{app:ggp} Figure~\ref{fig:jsd_persona_prompt_difference}) we use the JSON-format for all following experiments, expecting the results to generalize to other persona formats. However, we make both the JSON as well as the full-text version of the \emph{GGSS Personas} publicly available. We validate the full-text personas that are LLM-generated from the JSON-personas as input. Two co-authors of this paper independently annotated a sample of $80$ personas ($20$ randomly-sampled personas per TOP-$2$, -$4$, -$8$ and -$16$ collection) for hallucinations, misrepresentation or omission of individual attributes. We find that $69$ of $80$ (or $86.25$\%) full-text personas are accurate depictions of the JSON-personas, with errors occurring rarely through misrepresentation, e.g., of double negations or complex temporal orders, or through the omission of single attributes. We further describe the human validation process in Appendix~\ref{app:human_validation}.

\section{Preliminary Analysis}
\label{sec:empirical}

\subsection{Experimental Setup}
\label{sec:exp_setup}

Possible use cases and tasks of representative persona collections are plentiful. In this paper we focus on evaluating the utility of \emph{GGSS Personas} for simulating the response distribution of survey participants across a diverse set of outcome variables. 
% We split the 5246 ALLBUS respondents in training (50\%) and test respondents (50\%) and predict the response distribution of test respondents for selected survey variables. 

\paragraph{Outcome Variables}
We randomly sample $27$ survey variables across a range of different topics\textemdash three variables from each of the nine different topic areas covered in the ALLBUS\textemdash serving as outcome variables for which we simulate response distributions. We use the persona prompts as input and instruct the model to predict the response of the corresponding individual for the outcome variable. As described above, the outcome variables have been excluded from the set of persona attributes used to construct the persona prompts. 
% Where this occurs, we fill the set of top-$k$ attributes by including the next most important variable according to our rating.

For the 2023 version of the ALLBUS, the \textit{key topics} defined in the documentation are: \textit{Lifestyle}, \textit{Social Inequality}, \textit{Religion}, \textit{Ethnocentrism}, and \textit{Political Tendency}. We construct four additional topics from the large set of remaining variables and their categorization: \textit{Values \& Life Goals}, \textit{Economic Situation}, \textit{Social Capital}, and \textit{Morality}. Appendix~\ref{app:ggp} Table~\ref{tab:task_variables} provides details on the $27$ sampled outcome variables.  

% We split the 5246 ALLBUS respondents in training (50\%) and test respondents (50\%). We predict the distribution of the 27 selected survey variables for the test respondents.  We systematically vary the number of attributes $k$ made available to the baselines and the LLM, as well as the number of training samples $n$ (i.e., we systematically decrease the number of training respondents that the baseline classifiers have access to). The LLMs have no access to training respondents and can therefore also be used when no survey data is available.

\paragraph{Model Selection}
For generalizability, we selected five open weights, instruction-tuned LLMs, varying in size, developer, and origin. For some results, we average the performance of \textsc{Mistral-7B}, \textsc{Llama-3.1-8B}, and \textsc{Qwen3-8B}, referring to them as the \emph{7/8B LLMs} models. In comparison, we consider \textsc{Gemma-3-12B-it} and \textsc{Llama-3.3-70B} to be medium and large models, respectively.

We document the release and knowledge cutoff dates of the selected LLMs (see Appendix~\ref{app:ex_setup} Table~\ref{tab:model_releases})
showing that there is no issue with potential data leakage or contamination of the training data. All known knowledge cutoff dates fall before the first release of the ALLBUScompact (10/01/2025). The only model with an unknown knowledge cut-off date released after the release of the ALLBUScompact is \textsc{Qwen3-8B}, released 27/04/2025. However, given the usual gap of multiple months between the beginning of its training process (the relevant point in time for the knowledge cutoff) and the release of a model, we can be (almost) certain that none of the models we are using in this work could have had access to any of the ALLBUS data during training. 
% In addition, \citet{cheng2024dated} have shown that the effective knowledge cut-off date is even earlier than the official knowledge cutoff date, the date at which the last bit of training data was gathered. 

\paragraph{Baseline Method}
As baselines for the survey response distribution prediction task, we fit random forest classifiers on training datasets of varying size and with varying numbers of persona attributes used as input features. 
% The random forest classifier is a long-established method for data imputation, particularly when dealing with mixed data types and when missingness of data is a concern. It is still highly popular for researchers across different discipline \cite{adnan2022review}, able to match the performance of more advanced methods \cite{jager2021benchmark}, and particularly praised for its efficiency in complex settings with small sample size and high numbers of features with complex interactions \cite{dagdoug2023imputation}. 
We use the predictions of the random forest on an individual level and aggregate them across all personas to generate a simulated survey response distribution. 

This choice of baseline puts the persona-based zero-shot LLM approach at a disadvantage. While our approach always acts on a single persona description as input, the random forest classifiers are fitted on training samples representing persona descriptions and their corresponding responses for the respective prediction tasks, with training dataset sizes systematically ranging from $n=2$ to $n=2,048$. We reserve $20$\% of respondents as fixed test set, and randomly select $n$ of the remaining respondents for training the baseline classifiers. The setting of classifier hyperparameters follows \citet{miranda2025simulating} and is reported in Appendix~\ref{app:evaluation}. 
We compare the classifier predictions with the groundtruth response distribution of the test set respondents. The LLMs do not have access to any training samples.  

Similarly, we vary the number of persona attributes $k$ used as input features, systematically ranging from $k=2$ to $k=380$. For every $k$, we construct the set of persona attributes by selecting the $k$ most important attributes according to the variable importance ranking presented in Section~\ref{sec:ggp}.

\paragraph{Evaluation}
We measure the performance of our approach as well as of the different baselines by calculating the Jensen-Shannon Distance (JSD) between the predicted and the actual response distribution of each sampled outcome variable. The predicted response distribution is created by aggregating the model predictions for a given variable across all participants, and the actual response distribution is the aggregate of the participants' responses found in the ALLBUS. This procedure is in line with the suggestion offered by \citet{sorensen2024position} for the evaluation of distributionally pluralistic models.

Thus, the JSD measures how well the predicted responses approximate the actual responses across the entire range of different survey participants. The JSD is normalized to the range from $0$ to $1$, with lower values indicating higher similarity between the two distributions and thus better performance of the model in predicting the response distribution. We report the JSD for either the individual variables selected as prediction tasks or averaged across the full set of $27$ prediction task variables. 

\paragraph{Population Selection}
As discussed above, the \emph{GGSS Personas} is representative of the German population because it mirrors the randomly sampled participants of the ALLBUS.

In the experiments in which we are interested in the effects of representativity on the ability of persona-prompted LLMs to replicate the population's response distribution, we systematically create persona collections that are by design \emph{un}representative, e.g., oversampled on specific attributes. We sample $500$ participants from the highest income class to oversample on \textit{income}, $500$ participants that lean conservative (based on their responses to the TOP-$2$ attributes, see Figure~\ref{fig:figure_1}) to oversample on ideological leaning, and $500$ students to mimic a typical convenience sample population in social science studies \cite{sears1986college}.

% For each of the core demographic attributes \textit{gender}, \textit{income}, \textit{education}, and \textit{eastwest}, we create populations that include exclusively individuals from the extremes of the respective value space (oversampling).  We achieve this by randomly sampling $500$ individuals from these extremes. For the binary (\textit{eastwest}) and ordinal (\textit{education}) variables, we sample from the two (most extreme) available classes. For the categorical variable (\textit{gender}), we create non-representative populations for the classes \textit{male} and \textit{female}, each containing only \textit{males} or \textit{females}. For income we sample in total $500$ individuals with lowest and highest values, respectively. In Figure~\ref{fig:Figure5_baselines} we present the over-sampled students, high-income (> 10.000€/month) and overly conservative (extreme values for top TOP-2 attributes, seen in Figure~\ref{fig:figure_1}) populations. 

In addition, we run the same survey response tasks with the popular, but empirically not grounded \textit{PersonaHub} collection \cite{ge2024scaling}, as well as using \textit{no personas} at all. To ensure comparability of the results, we randomly sampled a subset of $500$ personas from the representative \emph{GGSS Personas}. By including multiple baselines we try to identify whether \textit{GGSS Personas}'s representativeness offers any advantages over non-representative alternatives.

% To ensure robustness, we repeat this procedure $5$ times for each of the unrepresentative populations, reporting the average results across the independent runs. 
\begin{figure}[t]
    \centering
    \includegraphics[width=\linewidth]{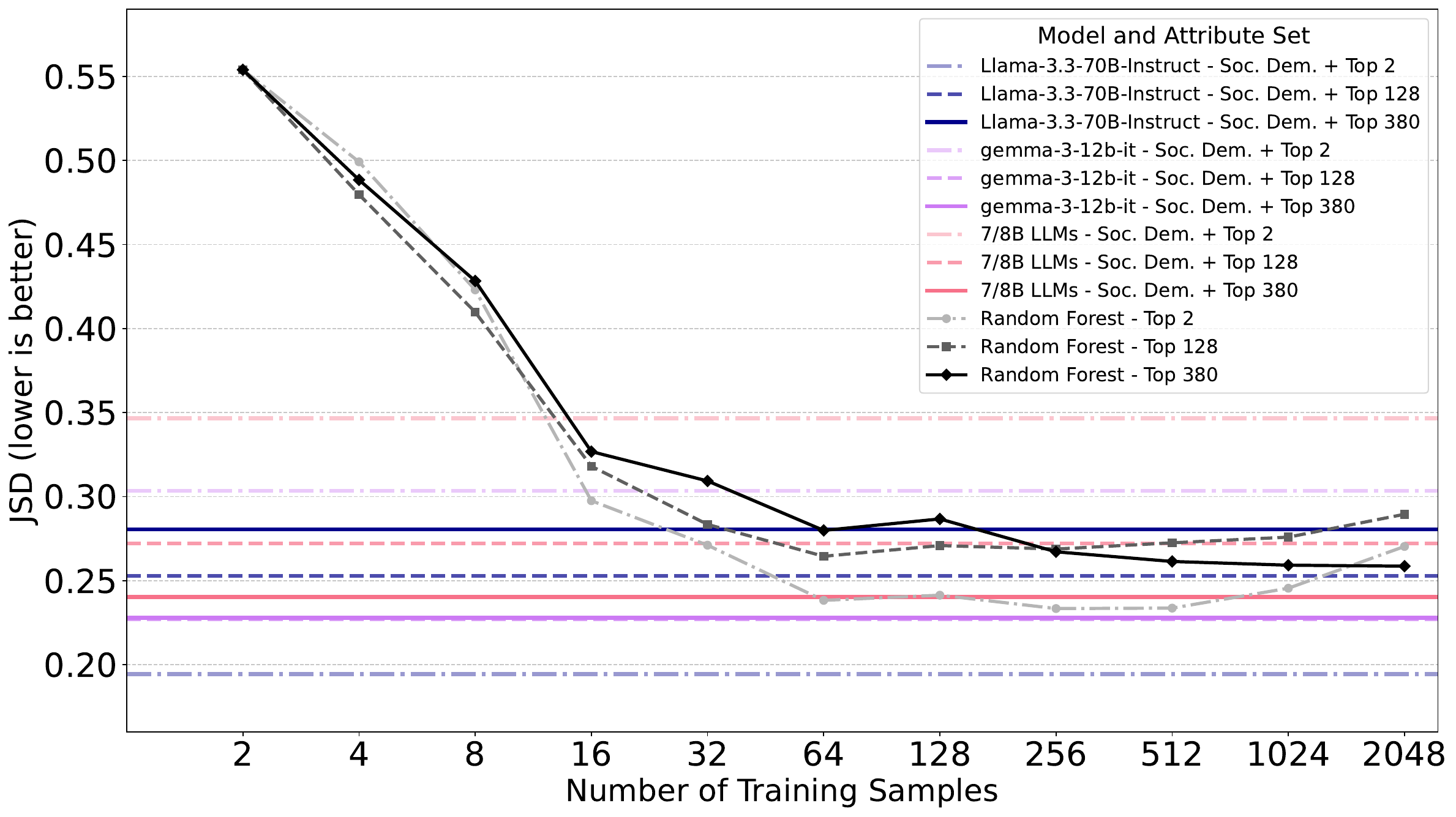}
    \caption{\textbf{Change in alignment with increased number of training samples.} We show the JSD between the survey response distribution and response distributions generated using persona-prompting with different LLMs as well as response distributions produced using random forest classifiers with increasing training set sizes. The alignment (averaged across $27$ outcome variables) is better when using persona-prompting, particularly for small $n$\textemdash LLMs are already well-aligned, even without training samples.}
    % \caption{\textbf{Zero-shot persona-based LLMs versus random forest baselines.} Performance (averaged JSD over 27 variables) is evaluated varying attribute count ($k$) and baseline training samples ($n$, LLMs are zero-shot). LLMs outperform random forests for small $n$. Medium and large LLMs are already well aligned without training examples, and for \textsc{Llama-3.3-70B-Instruct}, fewer persona attributes ($k$) show better alignment.}
    \label{fig:sparse_data}
\end{figure}

\paragraph{Response Generation}
\vspace*{-0.9\baselineskip}
We generate the synthetic survey response distribution for each of the selected tasks by prompting the LLMs with the persona descriptions featuring the socio-demographic characteristics and the TOP-$k$ attributes of each of the $5,246$ ALLBUS participants. The models are restricted to generate only the first token of the available responses, thus indicating the corresponding response label. We use \textit{vLLM} without varying any of the default response generation parameters. An exemplary prompt is given in Appendix~\ref{app:ex_setup}.

% To generate synthetic survey responses on the selected tasks, we prompt each model $5,246$ times per survey response prediction task. Each prompt reflects the persona attributes of an ALLBUS persona, with regard to its sociodemographic characteristics and the selected $top-k$ attributes. An exemplary prompt is given in Appendix~\ref{app:ex_setup}.

To generate responses for the \textit{PersonaHub} collection, we include the full-text persona description in the same manner. We only prompt the model with the survey question of the prediction task to generate responses with \textit{no persona}.

We only observe minor difficulties with models refusing to generate valid responses, with shares of invalid responses ranging from $1.44$\% for \textsc{Llama-3.1-8B} to $9.91$\% for \textsc{Qwen3-8B} (see Appendix~\ref{app:ex_setup} Table~\ref{tab:missings_topk_personas_percentage}). 
Interestingly, the share of invalid responses increases when the \textit{PersonaHub} or the oversampled populations are used.

\begin{figure}
    \centering
    \includegraphics[width=\linewidth]{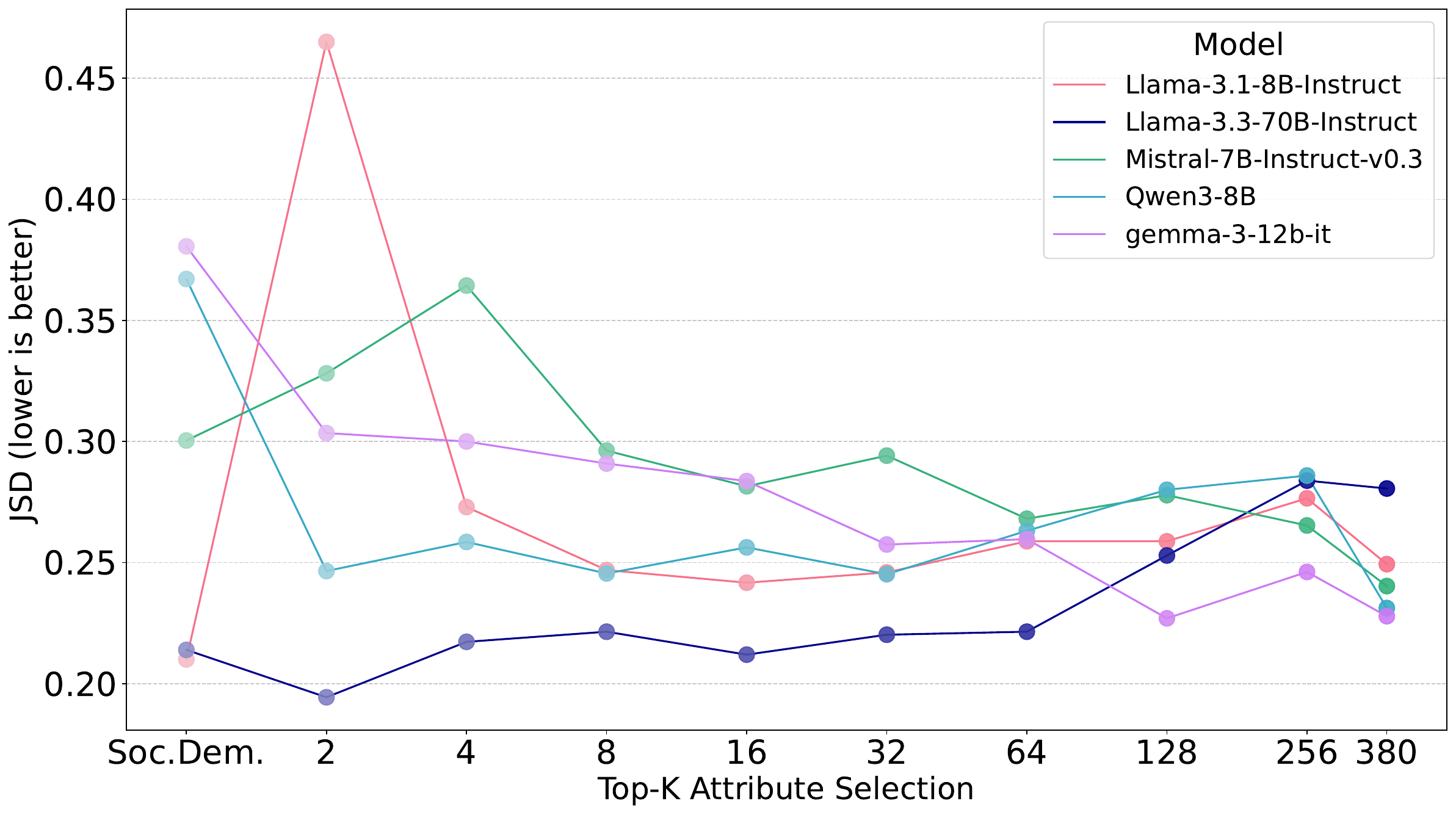}
    \caption{\textbf{Change in alignment with increased number of persona attributes.} We show the JSD between the survey response distribution and response distributions generated using increasingly large sets of TOP-$k$ persona attributes for different LLMs. The largest \textsc{Llama} model outperforms others up until the TOP-$64$ attributes are used, showing best alignment when using only the TOP-$2$ attributes. Across models, adding persona attributes does not monotonously lead to better alignment.
    % \textbf{Change in alignment with increased amount of persona attributes.} We exponentially increase the amount of \textit{k} persona attributes to investigate its impact on the alignment performance. Additional persona attributes does not necessarily improve alignment and remains rather unchanged after including the top 17\% of most important attributes ($k = 16$). The largest \textsc{Llama} model outperforms all other tested LLMs in areas where only up to 64 of the most relevant variables are available showing the most aligned responses with only the two most important attributes. After that point additional attributes seem to introduce noise and confusion worsens the alignment.
    }
    \label{fig:attributes}
\end{figure}
\vspace*{-0.9\baselineskip}
\subsection{Results}

% In this section, we compare the performance of different LLMs leveraging the \emph{GGSS Personas} in predicting the response distributions of a set of outcome variables against different random forest baselines, representing well-established alternatives for predicting survey responses.

We show that persona-prompting with the \emph{GGSS Personas} offers advantages over traditional statistical methods for survey response predictions in situations marked by data sparsity with few survey participants and little information about them available (Figure~\ref{fig:sparse_data}), and that increasing the level of information reflected in persona descriptions does not necessarily lead to improved performance (Figure~\ref{fig:attributes}).

\begin{figure}[t]
    \centering
    \includegraphics[width=\columnwidth]{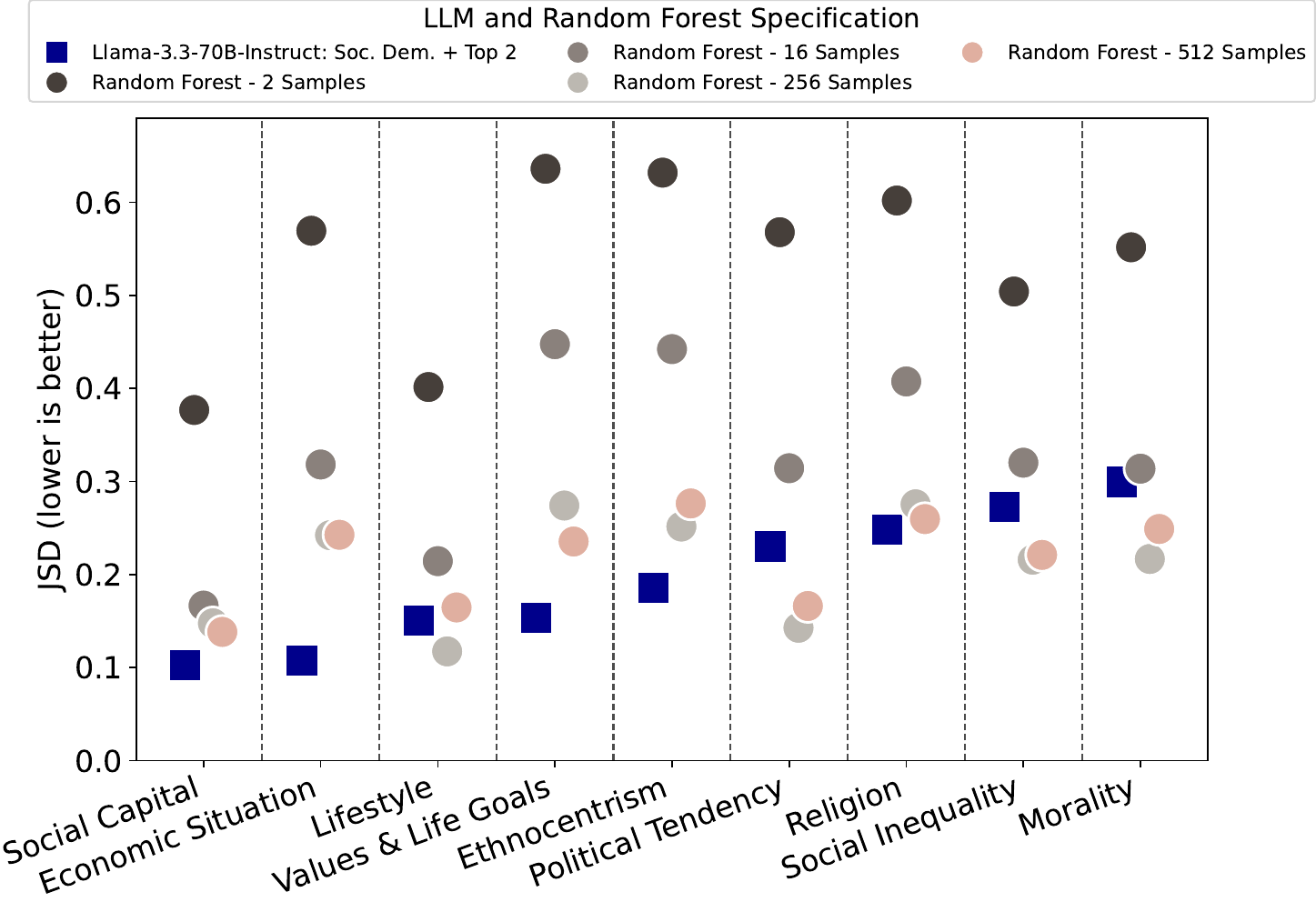}
    \caption{\textbf{Alignment comparison across different topics.} We compare the Jensen-Shannon Distance (JSD) between the survey response distribution and the response distribution generated using the best \emph{GGSS Personas} configuration as well as the response distributions from random forest baselines with different training set sizes. Using the \emph{GGSS Personas} and thus no input other than a single persona description produces response distributions that are better aligned with survey responses than the best random forest baselines across five out of nine different topics.}
    % \caption{\textbf{Alignment comparison of the best zero-shot persona-based LLM approach with random forest baselines across different thematic categories.} The performance is measured by the average Jensen-Shannon Distance (JSD) between the aggregated predicted and actual response distributions per variable. Our approach using no input other than the persona description outperforms the best performing random forest baselines for 13 out of 27 variables and in two of the nine topics with the task.}
    \label{fig:Figure4_jsd_tasks}
\end{figure}

\begin{figure}[t]
    \centering
    \includegraphics[width=\columnwidth]{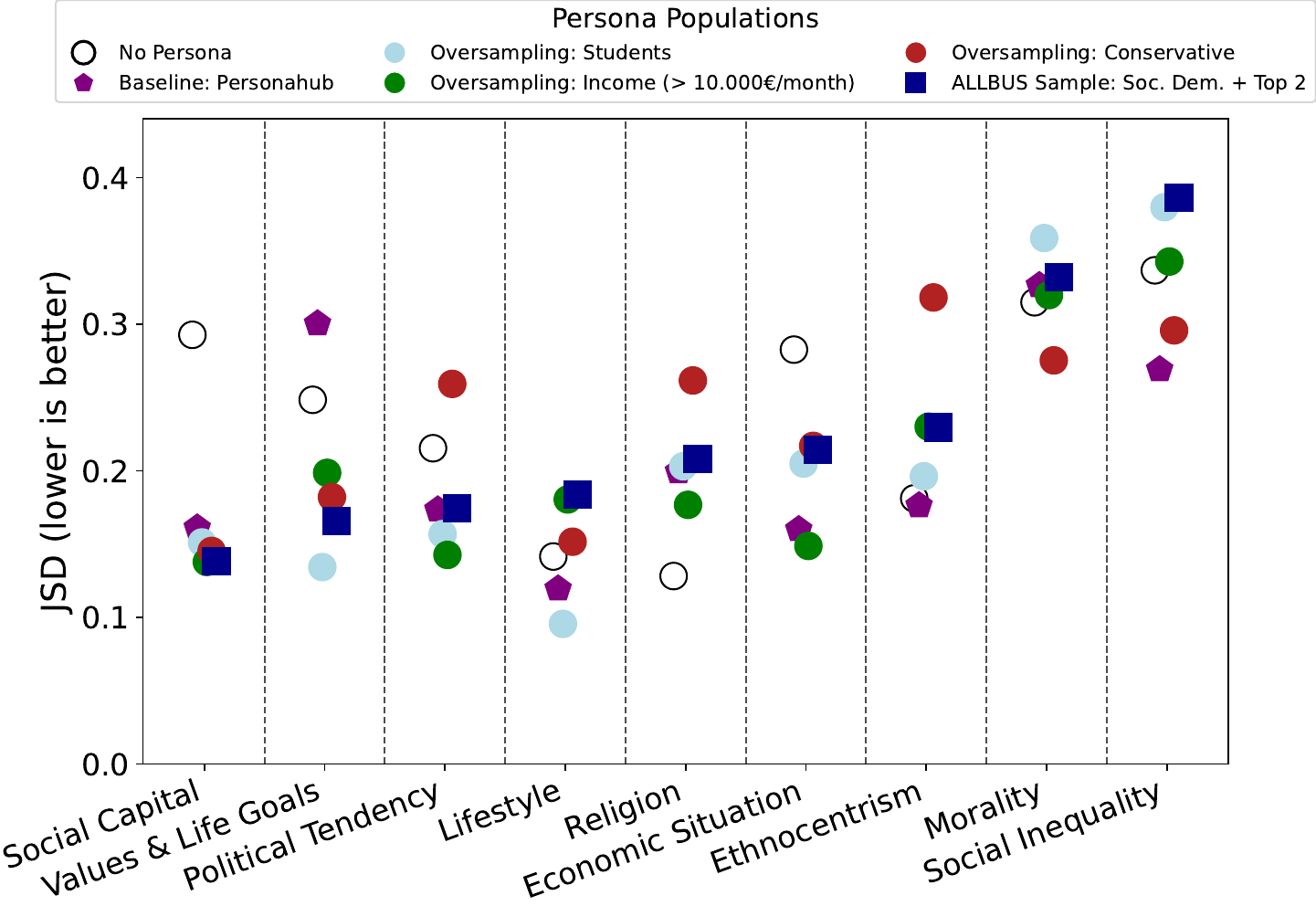}
    \caption{\textbf{Alignment comparison with unrepresentative persona collections.} We compare the average Jensen-Shannon Distance of response distributions generated with representative (\emph{GGSS Personas}), unrepresentative (baselines), and \textit{no persona} collections to the survey response distribution. All synthetic responses are generated using \textsc{Llama-3.3-70B-Instruct}. The unrepresentative baselines and the \emph{GGSS Personas} cluster closely together in terms of their JSD scores, indicating that representativity only has a small influence on the alignment.}
    \label{fig:Figure5_baselines}
\end{figure}

Additionally, we show that using the \emph{GGSS Personas} leads to predicted response distributions that are better aligned with the survey response distribution than those produced using established baselines in $13$ out of $27$ tasks and across five out of nine topics (Figure~\ref{fig:Figure4_jsd_tasks}).
Finally, we show that the representativity of the \emph{GGSS Personas} seems to have only little influence on the level of alignment (Figure~\ref{fig:Figure5_baselines}).

% Finally, we establish the potential for exciting further improvements of our approach by testing the effects of including training samples in the generation prompts (moving from zero- to few-shot prompting) and of increasing the size of the base LLM (Figure \ref{fig:mock_models}).

% \paragraph{RQ1 - Using persona prompts to steer zero-shot LLM predictions outperforms random forest predictions.}

\paragraph{Alignment for varying data availability.} A main advantage of the persona-prompting approach is its ability to tap the world-knowledge of LLMs, allowing them to adapt without much input data to diverse new contexts. Thus, we are interested in evaluating how our approach fares in situations of data scarcity. We simulate these situations by varying the number of training samples $n$ made available to the random forest classifiers serving as baselines, and by considering different sets of persona attributes $k$ included in the persona prompts and passed as input to the baseline classifiers. 

Figure~\ref{fig:sparse_data} shows that LLM predictions of the survey response distribution are generally well-aligned, even though the LLM does not have access to any training data. Compared with the random forest classifiers trained on increasingly large training sets, the best performing LLMs outperform these baselines across all constellations. Generally, the persona-prompted LLMs offer the greatest advantage over the established alternative for survey response prediction when training data is scarce, i.e., when the random forest classifiers are trained on only small numbers of training samples. Figure~\ref{fig:sparse_data} already indicates that using more persona attributes in persona prompts does not necessarily lead to better alignment with the survey response distribution, as shown by the fact that the best alignment results when using the largest model \textsc{Llama-3.3-70B-Instruct}, but only the TOP-$2$ attributes for constructing the personas.

We examine this observation more closely in Figure~\ref{fig:attributes}, which shows that there is no monotonous relationship between the number of TOP-$k$ attributes used for constructing the personas and the resulting alignment between the generated and the actual response distributions. Surprisingly, it is again the version of the \emph{GGSS Personas} that uses only the TOP-$2$ persona attributes that leads to the best alignment across all constellations we tested. For the best-performing LLM, we even see a clear tendency for the alignment to degrade with higher numbers of included persona attributes, highlighting a potential signal-noise trade-off when including more and increasingly unimportant persona attributes. Figure~\ref{fig:attributes} also reveals the general tendency that the best possible alignment improves with the size of the persona-prompted LLM used for generating the synthetic response distribution. Only the unlikely strong alignment of the relatively small \textsc{Llama-3.1-8B-Instruct} prompted with only socio-demographic information breaks this trend, warranting closer investigation. 

% In Figure~\ref{fig:sparse_data} we identified that the classifier with only $k=2$ attributes performs after being fitted with 32 or more training samples compared to the other classifier specifications. Additional attributes, degrading in importance with higher $k$ as they were ranked by feature importance, do not improve the alignment of the classifier predictions. This might highlight a potential noise signal trade-off when including increasingly more unimportant attributes.

% After that point, the classifier using the full set of $k=380$ persona attributes outperforms the other baselines.

\paragraph{Alignment across topics and survey questions.} In Figure~\ref{fig:Figure4_jsd_tasks}, we compare the performance of persona-prompting LLMs using the \emph{GGSS Personas} against random forest classifiers fitted on increasing numbers of training samples, now showing the disaggregated Jensen-Shannon Distances measured across nine different topics. Based on Figure~\ref{fig:attributes}, we use the best performing constellation for predicting response distributions, i.e., \textsc{Llama-3.3-70B-Instruct} prompted with TOP-$2$ personas, and increase the number of training samples for the baseline classifiers up to $n=512$, after which we do not observe any more performance improvements. We report the average JSD between the predicted and actual survey response distributions across the three outcome variables randomly sampled for each topic. 

% In Figure~\ref{fig:attributes} we showed that personas informed with the sociodemographic and the two most important ALLBUS variables reached the best alignment with the ground truth on average with \textsc{Llama-3.3-70B-Instruct}.

% Do we want to leave the references in there?

% We report the results for all other models, the number of attributes $k$ that are made available to the LLM and to the baseline Random Forest Classifier ($k=2$)  in the Appendix (\todo{add reference to appendix}). 

Our persona-prompted LLM, receiving only the TOP-$2$ persona attributes in the prompt, outperforms even the best informed baseline with access to up to $n=512$ of training samples and the same set of attributes in predicting the response distribution for $13$ out of $27$ different outcome variables. On the topic-level, the LLM-based approach turns out to be particularly strong in predicting response distributions in the categories \textit{Social Capital}, \textit{Economic Situation}, \textit{Religion}, \textit{Value \& Life Goals}, and \textit{Ethnocentrism}, where it achieves lower average JSD to the actual response distribution than even the best random forest classifiers. The absolute alignment using the \emph{GGSS Personas} to persona-prompt LLMs is best for the topics \textit{Social Capital} and \textit{Economic Situation}, and worst for the topics \textit{Social Inequality} and \textit{Morality}. 

Generally, the strong performance of the persona-prompt approach for at least some topics should be considered in the context of the fact that the baseline classifiers were able to learn persona-response-patterns from the training data, while the persona-based approach is purely zero-shot. The LLM has to rely solely on its \textit{world knowledge} and the associations triggered by the single persona description.

% For categories \textit{value and life goals}, \textit{ethnocentrism}, \textit{lifestyle} and \textit{religion}, our approach outperforms the best baseline for two out of three variables, and for \textit{political tendency} and \textit{social capital} for one out of three variables. 

% From Figure~\ref{fig:Figure4_jsd_tasks}, we thus take that our persona-based approach is able to deliver comparable if not better performance across a range of variables and thematic categories. Most remarkably, our approach achieves this performance with only a single persona description corresponding to the individual whose response is to be predicted. The random forest classifier instead, does not only have access to this individual's persona description, but to the persona descriptions and responses of many other individuals in the training set. While the baseline classifier was able to learn persona-response-patterns from this data, the persona-based LLM approach had to rely on its \textit{world knowledge} and the associations triggered by the single persona description.

\paragraph{Alignment compared to baseline persona populations.}
We run the same experiments using unrepresentative\textemdash oversampled on specific attributes and not empirically grounded\textemdash persona collections as baselines, as described in Section~\ref{sec:exp_setup}. In this comparison, there is no clear pattern showing that response distributions predicted using the representative \emph{GGSS Personas} are consistently better aligned than the baselines they are compared against. Instead, Figure~\ref{fig:Figure5_baselines} shows that the different persona collections are generally scattered closely together around similar alignment scores across different topics, indicating that the predictive performance is highly topic dependent. 

For the topics \textit{Social Capital}, \textit{Values \& Life Goals}, \textit{Political Tendency} and \textit{Economic Situation}, the persona-prompted response distributions are closer aligned with the survey response distribution than the response distributions generated using \textit{no personas}, indicating that the TOP-$2$ attributes used in the persona descriptions might have been helpful in correctly predicting the responses in these areas. The baseline oversampled on conservatives (as derived from the TOP-$2$ attributes) deviates most strongly from the \emph{GGSS Personas}, strengthening our assumption that these attributes have the highest impact on the prediction of other ALLBUS variables. The fact that the empirically ungrounded \textit{PersonaHub} collection aligns so closely with the different persona collections based on the ALLBUS is surprising.    

% Compared to the classifier results, we observe a different outcome in Figure~\ref{fig:Figure5_baselines} as predictions based on the representative \emph{GGSS Personas} are not consistently better aligned than baseline population. In general, the results hint that persona prompting could improve alignment compared to prompting without personas, e.g. in \textit{social capital}, \textit{economic situation}, \textit{political tendency} and \textit{value \& life goals}. 

% However, the results are highly topic-dependent. There seem to be topics that are harder to grasp with persona prompting, as most of the persona populations are scattered around similar alignment scores without many outliers.

\section{Discussion}
\label{sec:discussion}

This work introduces \emph{German General Social Survey Personas} - a novel prompt collection designed to evaluate and enhance the use of persona-based prompting for modeling response distributions of populations. Our preliminary experiments demonstrate the current ability of persona-based approaches to outperform traditional response prediction and imputation methods, particularly in situations characterized by limited data availability. These findings suggest that persona prompting can serve as a viable and flexible alternative in low-resource contexts where conventional supervised methods may struggle.

A key consideration in developing \emph{German General Social Survey Personas} is achieving a balance between generalizability\textemdash ensuring that the collection remains applicable across a wide range of downstream tasks\textemdash and specificity, which is essential for strong performance in domain-specific applications. In this work, we propose leveraging general population surveys to construct empirically informed persona collections. Interestingly, our results indicate that personas based on a small subset of key variables outperform those that incorporate all available survey variables. This suggests that including only the most informative attributes may enhance both efficiency and predictive accuracy. In our preliminary analysis we only considered the average importance of variables for all other outcome variable. An important direction for future research is to explore how persona descriptions can be systematically tuned to optimally align with observed response patterns.

Recent work has begun fine-tuning large language models (LLMs) not only to generate single “correct” answers, but to minimize the divergence between the model’s predicted response distributions and empirical human survey data from sources such as the World Values Survey and Pew Global Attitudes Survey \citep{cao2025specializing, suh2025language}. However, these supervised distribution-fitting methods require substantial computational resources and access to proprietary model weights, making them less applicable in low-resource or restricted-access settings. In contrast, persona prompting offers a more lightweight and adaptable alternative that can be applied even to publicly available models. The performance of persona-based methods can potentially further be improved through few-shot approaches and with the release of increasingly capable language models. This highlights a promising avenue for future work, combining the interpretability and flexibility of prompting with the precision gains typically associated with fine-tuned models. 

\emph{German General Social Survey Personas} can act as a valuable resource that supports systematic comparisons of population-alignment of LLMs within different tasks (e.g., survey response generation, discussion simulations or behavior predictions). 
\section{Limitations}
\label{sec:limitations}

The idea of turning general social surveys such as the ALLBUS into empirically grounded persona collection comes with a number of limitations. First, surveys only offer a \textbf{snapshot of a population that dynamically changes over time}, both in its composition (e.g., due to migration or demographic change) as well as in the attitudes and beliefs held by individuals. To explicitly mark this temporal dependency, we date the released \emph{GGSS Personas} through versioning. A partial alleviation of this limitation could also be in the use of reweighting techniques, which we leave for future work to explore. 

Second, while we rule out the possibility that any of our tested LLMs has had access to the ALLBUS data we use, another (indirect) form of \textbf{data leakage} could occur across different releases of the ALLBUS, given that such surveys to a large degree run the same questions in every iteration. Future work could thus investigate performance differences between \emph{GGSS Personas} versions created from and evaluated on current and past releases of the ALLBUS. 

Third, missing information at an individual level introduces a \textbf{trade-off between the representativity of the persona collection and the completeness of the individual persona descriptions}. In our experiments, we prioritize representativity by including all personas instead of selecting only those that are complete, i.e., have information on all attributes available. By releasing the \emph{GGSS Personas} with all personas, including those that are incomplete, we allow users of the collection to resolve this trade-off as they see fit\textemdash they could prioritize representativity by working with the collection as is, or they could prioritize persona completeness by selecting only personas with full information. Future work could explore how imputation procedures, possibly LLM-based \cite{castricato2025persona}, might help to resolve this trade-off. 

Fourth, our \textbf{attribute selection procedure does not account for correlations} and possible multicollinearity between predictors. Future work could test methods that evaluate all possible subsets of $k$ attributes to improve upon the sets of TOP-$k$ attributes currently used for persona creation.

In addition, LLM-generated closed-ended survey responses are susceptible to minimal perturbations, such as the positioning of the answer option \cite{rupprecht2025promptperturbationsrevealhumanlike, tjuatja_llms_2024}. Special attention should be put on the robustness of synthetic survey responses by applying prompt perturbations to ensure that the results are not only artifacts of the prompting style.

Lastly, we are \textbf{relying exclusively on survey data}, both for the creation of persona descriptions and as a source of human groundtruth to evaluate our approach against. However, there is some evidence that survey data itself might be limited in accurately capturing human attitudes and behaviors. Future work could on the one hand explore the inclusion of behavioral data in persona prompts as a more immediate proxy of human behavior, and on the other hand search additional sources of groundtruth (such as election outcomes) to evaluate against.

\section*{Ethical Considerations}

Generating artificial personas and representative collections might be relevant in various domains and applied to different use cases. However, surveying an artificial persona collection instead of the real, underlying population can be senseless or even dangerous depending on the downstream task at hand. 
For example, artificial persona collections can be exploited to pre-test and optimize targeted political or manipulative messaging. Such use risks amplifying disinformation campaigns and undermining democratic processes. Further, over-reliance on persona collections and its results is risky when there is no ground truth data of the real target population available as the alignment of the artificial responses cannot be evaluated. Frequent reliance on artificial responses may normalize their use where human perspectives are irreplaceable (e.g. in policymaking or clinical trials). This risks sidelining real human voices in domains directly impacting human lives.
Researchers should also consider ethical evasion as one possible issue with synthetic persona collections and responses. Synthetic respondents might be viewed like a way to bypass obligatory ethical review processes since no real human participants are involved. This might encourage under-regulated research practices and in the long run weaken ethical safeguards.

\section*{Acknowledgement}
We thank all anonymous reviewers for their thoughtful feedback and constructive suggestions, which greatly improved the quality of this work. The authors acknowledge support by the state of Baden-Württemberg through bwHPC and the German Research Foundation (DFG) through grant INST 35/1597-1 FUGG.
% \nocite{*}
\section{References}\label{sec:reference}

\bibliographystyle{lrec2026-natbib}
\bibliography{references}

@inproceedings{aher2023using,
  title        = {{Using Large Language Models to Simulate Multiple Humans and Replicate Human Subject Studies}},
  author       = {Aher, Gati V. and Arriaga, Rosa I. and Kalai, Adam Tauman},
  booktitle    = {Proceedings of the 40th International Conference on Machine Learning},
  pages        = {337--371},
  year         = {2023},
  editor       = {Krause, Andreas and Brunskill, Emma and Cho, Kyunghyun and Engelhardt, Barbara and Sabato, Sivan and Scarlett, Jonathan},
  volume       = {202},
  series       = {Proceedings of Machine Learning Research},
  publisher    = {PMLR}
}

@misc{allbus2023,
  author        = {GESIS Leibniz-Institut f{\"u}r Sozialwissenschaften},
  title         = {{German General Social Survey ALLBUS 2023}},
  year          = {2025},
  howpublished  = {(ZA8830; Version 1.2.0) [Data set]. GESIS, Cologne. \url{https://doi.org/10.4232/1.14544}},
  doi           = {10.4232/1.14544}
}

@misc{allbus2023compact,
  author        = {GESIS Leibniz-Institut f{\"u}r Sozialwissenschaften},
  title         = {{German General Social Survey ALLBUScompact 2023}},
  year          = {2025},
  howpublished  = {(ZA8831; Version 1.3.0) [Data set]. GESIS, Cologne. \url{https://doi.org/10.4232/1.14545}},
  doi           = {10.4232/1.14545}
}

@inproceedings{anthis2025position,
  title        = {{Position: LLM Social Simulations Are a Promising Research Method}},
  author       = {Anthis, Jacy Reese and Liu, Ryan and Richardson, Sean M. and Kozlowski, Austin C. and Koch, Bernard and Brynjolfsson, Erik and Evans, James and Bernstein, Michael S.},
  booktitle    = {ICML 2025 Position Paper Track, 42nd International Conference on Machine Learning},
  year         = {2025}
}

@article{argyle2023out,
  title        = {{Out of One, Many: Using Language Models to Simulate Human Samples}},
  author       = {Argyle, Lisa P. and Busby, Ethan C. and Fulda, Nancy and Gubler, Joshua R. and Rytting, Christopher and Wingate, David},
  journal      = {Political Analysis},
  volume       = {31},
  number       = {3},
  pages        = {337--351},
  year         = {2023},
  publisher    = {Cambridge University Press}
}

@inproceedings{ashkinaze2025plurals,
  title        = {{Plurals: A System for Guiding LLMs via Simulated Social Ensembles}},
  author       = {Ashkinaze, Joshua and Fry, Emily and Edara, Narendra and Gilbert, Eric and Budak, Ceren},
  booktitle    = {Proceedings of the 2025 CHI Conference on Human Factors in Computing Systems},
  pages        = {1--21},
  year         = {2025}
}

@inproceedings{beck2024sensitivity,
  title        = {{Sensitivity, Performance, Robustness: Deconstructing the Effect of Sociodemographic Prompting}},
  author       = {Beck, Tilman and Schuff, Hendrik and Lauscher, Anne and Gurevych, Iryna},
  booktitle    = {Proceedings of the 18th Conference of the European Chapter of the Association for Computational Linguistics (Volume 1: Long Papers)},
  pages        = {2589--2615},
  year         = {2024}
}

@inproceedings{cao2025specializing,
  title        = {{Specializing Large Language Models to Simulate Survey Response Distributions for Global Populations}},
  author       = {Cao, Yong and Liu, Haijiang and Arora, Arnav and Augenstein, Isabelle and R{\"o}ttger, Paul and Hershcovich, Daniel},
  booktitle    = {Proceedings of the 2025 Conference of the Nations of the Americas Chapter of the Association for Computational Linguistics: Human Language Technologies (Volume 1: Long Papers)},
  pages        = {3141--3154},
  year         = {2025}
}

@inproceedings{castricato2025persona,
  title        = {{PERSONA: A Reproducible Testbed for Pluralistic Alignment}},
  author       = {Castricato, Louis and Lile, Nathan and Rafailov, Rafael and Fr{\"a}nken, Jan-Philipp and Finn, Chelsea},
  booktitle    = {Proceedings of the 31st International Conference on Computational Linguistics},
  pages        = {11348--11368},
  year         = {2025}
}

@article{chen2024from,
  title        = {{From Persona to Personalization: A Survey on Role-Playing Language Agents}},
  author       = {Jiangjie Chen and Xintao Wang and Rui Xu and Siyu Yuan and Yikai Zhang and Wei Shi and Jian Xie and Shuang Li and Ruihan Yang and Tinghui Zhu and Aili Chen and Nianqi Li and Lida Chen and Caiyu Hu and Siye Wu and Scott Ren and Ziquan Fu and Yanghua Xiao},
  journal      = {Transactions on Machine Learning Research},
  issn         = {2835-8856},
  year         = {2024},
  url          = {https://openreview.net/forum?id=xrO70E8UIZ},
  note         = {Survey Certification}
}

@inproceedings{chen2025personatwin,
  title        = {{PersonaTwin: A Multi-Tier Prompt Conditioning Framework for Generating and Evaluating Personalized Digital Twins}},
  author       = {Chen, Sihan and Lalor, John P. and Yang, Yi and Abbasi, Ahmed},
  booktitle    = {Proceedings of the Fourth Workshop on Generation, Evaluation and Metrics (GEM$^2$)},
  pages        = {774--788},
  year         = {2025}
}

@article{chen2025persona,
  title        = {{Persona Vectors: Monitoring and Controlling Character Traits in Language Models}},
  author       = {Chen, Runjin and Arditi, Andy and Sleight, Henry and Evans, Owain and Lindsey, Jack},
  journal      = {arXiv preprint arXiv:2507.21509},
  year         = {2025}
}

@inproceedings{cheng2023compost,
  title        = {{CoMPosT: Characterizing and Evaluating Caricature in LLM Simulations}},
  author       = {Cheng, Myra and Piccardi, Tiziano and Yang, Diyi},
  booktitle    = {Proceedings of the 2023 Conference on Empirical Methods in Natural Language Processing},
  pages        = {10853--10875},
  year         = {2023}
}

@inproceedings{cheng2023marked,
  title        = {{Marked Personas: Using Natural Language Prompts to Measure Stereotypes in Language Models}},
  author       = {Cheng, Myra and Durmus, Esin and Jurafsky, Dan},
  booktitle    = {Proceedings of the 61st Annual Meeting of the Association for Computational Linguistics (Volume 1: Long Papers)},
  pages        = {1504--1532},
  year         = {2023}
}

@article{de2025principled,
  title        = {{Principled Personas: Defining and Measuring the Intended Effects of Persona Prompting on Task Performance}},
  author       = {de Araujo, Pedro Henrique Luz and R{\"o}ttger, Paul and Hovy, Dirk and Roth, Benjamin},
  journal      = {arXiv preprint arXiv:2508.19764},
  year         = {2025}
}

@article{deng2025personateaming,
  title        = {{Personateaming: Exploring How Introducing Personas Can Improve Automated AI Red-Teaming}},
  author       = {Deng, Wesley Hanwen and Kim, Sunnie S. Y. and Jha, Akshita and Holstein, Ken and Eslami, Motahhare and Wilcox, Lauren and Gatys, Leon A.},
  journal      = {arXiv preprint arXiv:2509.03728},
  year         = {2025}
}

@inproceedings{durmus2024towards,
  title        = {{Towards Measuring the Representation of Subjective Global Opinions in Language Models}},
  author       = {Durmus, Esin and Nguyen, Karina and Liao, Thomas and Schiefer, Nicholas and Askell, Amanda and Bakhtin, Anton and Chen, Carol and Hatfield-Dodds, Zac and Hernandez, Danny and Joseph, Nicholas and others},
  booktitle    = {First Conference on Language Modeling},
  year         = {2024}
}

@inproceedings{feng2023pretraining,
  title        = {{From Pretraining Data to Language Models to Downstream Tasks: Tracking the Trails of Political Biases Leading to Unfair NLP Models}},
  author       = {Feng, Shangbin and Park, Chan Young and Liu, Yuhan and Tsvetkov, Yulia},
  booktitle    = {Proceedings of the 61st Annual Meeting of the Association for Computational Linguistics (Volume 1: Long Papers)},
  pages        = {11737--11762},
  year         = {2023}
}

@inproceedings{frohling2025personas,
  title        = {{Personas With Attitudes: Controlling LLMs for Diverse Data Annotation}},
  author       = {Fr{\"o}hling, Leon and Demartini, Gianluca and Assenmacher, Dennis},
  booktitle    = {Proceedings of the The 9th Workshop on Online Abuse and Harms (WOAH)},
  year         = {2025},
  url          = {https://aclanthology.org/2025.woah-1.43/}
}

@article{ge2024scaling,
  title        = {{Scaling Synthetic Data Creation With 1,000,000,000 Personas}},
  author       = {Ge, Tao and Chan, Xin and Wang, Xiaoyang and Yu, Dian and Mi, Haitao and Yu, Dong},
  journal      = {arXiv preprint arXiv:2406.20094},
  year         = {2024}
}

@article{gonzalez2025llms,
  title        = {{LLMs Model Non-WEIRD Populations: Experiments With Synthetic Cultural Agents}},
  author       = {Gonzalez-Bonorino, Augusto and Capra, Monica and Pantoja, Emilio},
  journal      = {arXiv preprint arXiv:2501.06834},
  year         = {2025}
}

@article{hewitt2024predicting,
  title        = {{Predicting Results of Social Science Experiments Using Large Language Models}},
  author       = {Hewitt, Luke and Ashokkumar, Ashwini and Ghezae, Isaias and Willer, Robb},
  journal      = {Preprint},
  year         = {2024}
}

@inproceedings{hu2024quantifying,
  title        = {{Quantifying the Persona Effect in LLM Simulations}},
  author       = {Hu, Tiancheng and Collier, Nigel},
  booktitle    = {Proceedings of the 62nd Annual Meeting of the Association for Computational Linguistics (Volume 1: Long Papers)},
  pages        = {10289--10307},
  year         = {2024}
}

@inproceedings{hwang2023aligning,
  title        = {{Aligning Language Models to User Opinions}},
  author       = {Hwang, Eunjeong and Majumder, Bodhisattwa and Tandon, Niket},
  booktitle    = {Findings of the Association for Computational Linguistics: EMNLP 2023},
  pages        = {5906--5919},
  year         = {2023}
}

@article{kim2023ai,
  title        = {{AI-Augmented Surveys: Leveraging Large Language Models and Surveys for Opinion Prediction}},
  author       = {Kim, Junsol and Lee, Byungkyu},
  journal      = {arXiv preprint arXiv:2305.09620},
  year         = {2023}
}

@article{kim2024persona,
  title        = {{Persona Is a Double-Edged Sword: Mitigating the Negative Impact of Role-Playing Prompts in Zero-Shot Reasoning Tasks}},
  author       = {Kim, Junseok and Yang, Nakyeong and Jung, Kyomin},
  journal      = {arXiv preprint arXiv:2408.08631},
  year         = {2024}
}

@article{kirk2024benefits,
  title        = {{The Benefits, Risks and Bounds of Personalizing the Alignment of Large Language Models to Individuals}},
  author       = {Kirk, Hannah Rose and Vidgen, Bertie and R{\"o}ttger, Paul and Hale, Scott A.},
  journal      = {Nature Machine Intelligence},
  volume       = {6},
  number       = {4},
  pages        = {383--392},
  year         = {2024},
  publisher    = {Nature Publishing Group UK London}
}

@article{kruskal1979representative,
  title        = {{Representative Sampling, II: Scientific Literature, Excluding Statistics}},
  author       = {Kruskal, William and Mosteller, Frederick},
  journal      = {International Statistical Review/Revue Internationale de Statistique},
  pages        = {111--127},
  year         = {1979},
  publisher    = {JSTOR}
}

@inproceedings{kwon2023efficient,
  title        = {{Efficient Memory Management for Large Language Model Serving With PagedAttention}},
  author       = {Kwon, Woosuk and Li, Zhuohan and Zhuang, Siyuan and Sheng, Ying and Zheng, Lianmin and Yu, Cody Hao and Gonzalez, Joseph and Zhang, Hao and Stoica, Ion},
  booktitle    = {Proceedings of the 29th Symposium on Operating Systems Principles},
  pages        = {611--626},
  year         = {2023}
}

@article{lutz2025prompt,
  title        = {{The Prompt Makes the Person(a): A Systematic Evaluation of Sociodemographic Persona Prompting for Large Language Models}},
  author       = {Lutz, Marlene and Sen, Indira and Ahnert, Georg and Rogers, Elisa and Strohmaier, Markus},
  journal      = {arXiv preprint arXiv:2507.16076},
  year         = {2025}
}

@inproceedings{ma2025algorithmic,
  title        = {{Algorithmic Fidelity of Large Language Models in Generating Synthetic German Public Opinions: A Case Study}},
  author       = {Ma, Bolei and Yoztyurk, Berk and Haensch, Anna-Carolina and Wang, Xinpeng and Herklotz, Markus and Kreuter, Frauke and Plank, Barbara and A{\ss}enmacher, Matthias},
  booktitle    = {Proceedings of the 63rd Annual Meeting of the Association for Computational Linguistics (Volume 1: Long Papers)},
  year         = {2025},
  url          = {https://aclanthology.org/2025.acl-long.90/},
  doi          = {10.18653/v1/2025.acl-long.90},
  pages        = {1785--1809}
}

@article{miranda2025simulating,
  title        = {{Simulating Public Opinion: Comparing Distributional and Individual-Level Predictions from LLMs and Random Forests}},
  author       = {Miranda, Fernando and Balbi, Pedro Paulo},
  journal      = {Entropy},
  volume       = {27},
  number       = {9},
  pages        = {923},
  year         = {2025},
  publisher    = {MDPI}
}

@article{orlikowski2025beyond,
  title        = {{Beyond Demographics: Fine-Tuning Large Language Models to Predict Individuals' Subjective Text Perceptions}},
  author       = {Orlikowski, Matthias and Pei, Jiaxin and R{\"o}ttger, Paul and Cimiano, Philipp and Jurgens, David and Hovy, Dirk},
  journal      = {arXiv preprint arXiv:2502.20897},
  year         = {2025}
}

@article{park2024generative,
  title        = {{Generative Agent Simulations of 1,000 People}},
  author       = {Park, Joon Sung and Zou, Carolyn Q. and Shaw, Aaron and Hill, Benjamin Mako and Cai, Carrie and Morris, Meredith Ringel and Willer, Robb and Liang, Percy and Bernstein, Michael S.},
  journal      = {arXiv preprint arXiv:2411.10109},
  year         = {2024}
}

@article{peng2025mega,
  title        = {{A Mega-Study of Digital Twins Reveals Strengths, Weaknesses and Opportunities for Further Improvement}},
  author       = {Peng, Tiany and Gui, George and Merlau, Daniel J. and Fan, Grace Jiarui and Sliman, Malek Ben and Brucks, Melanie and Johnson, Eric J. and Morwitz, Vicki and Althenayyan, Abdullah and Bellezza, Silvia and others},
  journal      = {arXiv preprint arXiv:2509.19088},
  year         = {2025}
}

@article{rupprecht2025promptperturbationsrevealhumanlike,
      title={Prompt Perturbations Reveal Human-Like Biases in Large Language Model Survey Responses}, 
      author={Jens Rupprecht and Georg Ahnert and Markus Strohmaier},
      journal      = {arXiv preprint arXiv:2507.07188},
      year         = {2025},
      archivePrefix={arXiv},
      primaryClass={cs.CL},
      url={https://arxiv.org/abs/2507.07188}, 
}

@article{salem2025tinytroupe,
  title        = {{TinyTroupe: An LLM-Powered Multiagent Persona Simulation Toolkit}},
  author       = {Salem, Paulo and Sim, Robert and Olsen, Christopher and Saxena, Prerit and Barcelos, Rafael and Ding, Yi},
  journal      = {arXiv preprint arXiv:2507.09788},
  year         = {2025}
}

@inproceedings{santurkar2023whose,
  title        = {{Whose Opinions Do Language Models Reflect?}},
  author       = {Santurkar, Shibani and Durmus, Esin and Ladhak, Faisal and Lee, Cinoo and Liang, Percy and Hashimoto, Tatsunori},
  booktitle    = {International Conference on Machine Learning},
  pages        = {29971--30004},
  year         = {2023},
  organization = {PMLR}
}

@article{sears1986college,
  title        = {{College Sophomores in the Laboratory: Influences of a Narrow Data Base on Social Psychology's View of Human Nature}},
  author       = {Sears, David O.},
  journal      = {Journal of Personality and Social Psychology},
  volume       = {51},
  number       = {3},
  pages        = {515},
  year         = {1986},
  publisher    = {American Psychological Association}
}

@inproceedings{sen2025missing,
  title        = {{Missing the Margins: A Systematic Literature Review on the Demographic Representativeness of LLMs}},
  booktitle    = {Findings of the Association for Computational Linguistics: ACL 2025},
  author       = {Sen, Indira and Lutz, Marlene and Rogers, Elisa and Garcia, David and Strohmaier, Markus},
  year         = {2025},
  pages        = {24263--24289},
  publisher    = {Association for Computational Linguistics},
  doi          = {10.18653/v1/2025.findings-acl.1246}
}

@inproceedings{shi2023large,
  title        = {{Large Language Models Can Be Easily Distracted by Irrelevant Context}},
  author       = {Shi, Freda and Chen, Xinyun and Misra, Kanishka and Scales, Nathan and Dohan, David and Chi, Ed H. and Sch{\"a}rli, Nathanael and Zhou, Denny},
  booktitle    = {International Conference on Machine Learning},
  pages        = {31210--31227},
  year         = {2023},
  organization = {PMLR}
}

@inproceedings{sorensen2024position,
  title        = {{Position: A Roadmap to Pluralistic Alignment}},
  author       = {Sorensen, Taylor and Moore, Jared and Fisher, Jillian and Gordon, Mitchell and Mireshghallah, Niloofar and Rytting, Christopher Michael and Ye, Andre and Jiang, Liwei and Lu, Ximing and Dziri, Nouha and others},
  booktitle    = {Proceedings of the 41st International Conference on Machine Learning},
  pages        = {46280--46302},
  year         = {2024}
}

@inproceedings{suh2025language,
  title        = {{Language Model Fine‑Tuning on Scaled Survey Data for Predicting Distributions of Public Opinions}},
  author       = {Suh, Joseph and Jahanparast, Erfan and Moon, Suhong and Kang, Minwoo and Chang, Serina},
  booktitle    = {80th Annual AAPOR Conference},
  year         = {2025},
  organization = {AAPOR}
}

@inproceedings{sun2025persona,
  title        = {{Persona‑DB: Efficient Large Language Model Personalization for Response Prediction With Collaborative Data Refinement}},
  author       = {Sun, Chenkai and Yang, Ke and Reddy, Revanth Gangi and Fung, Yi and Chan, Hou Pong and Small, Kevin and Zhai, ChengXiang and Ji, Heng},
  booktitle    = {Proceedings of the 31st International Conference on Computational Linguistics},
  pages        = {281--296},
  year         = {2025}
}

@article{toubia2025database,
  title        = {{Database Report: Twin‑2K‑500: A Data Set for Building Digital Twins of Over 2,000 People Based on Their Answers to Over 500 Questions}},
  author       = {Toubia, Olivier and Gui, George Z. and Peng, Tianyi and Merlau, Daniel J. and Li, Ang and Chen, Haozhe},
  journal      = {Marketing Science},
  year         = {2025},
  publisher    = {INFORMS}
}

@article{tjuatja_llms_2024,
	address = {Cambridge, MA},
	title = {Do {LLMs} {Exhibit} {Human}-like {Response} {Biases}? {A} {Case} {Study} in {Survey} {Design}},
	volume = {12},
	shorttitle = {Do {LLMs} {Exhibit} {Human}-like {Response} {Biases}?},
	url = {https://aclanthology.org/2024.tacl-1.56/},
	doi = {10.1162/tacl_a_00685},
	urldate = {2026-03-02},
	journal = {Transactions of the Association for Computational Linguistics},
	publisher = {MIT Press},
	author = {Tjuatja, Lindia and Chen, Valerie and Wu, Tongshuang and Talwalkwar, Ameet and Neubig, Graham},
	year = {2024},
	pages = {1011--1026},
}

@inproceedings{tseng2024two,
  title        = {{Two Tales of Persona in LLMs: A Survey of Role‑Playing and Personalization}},
  author       = {Tseng, Yu‑Min and Huang, Yu‑Chao and Hsiao, Teng‑Yun and Chen, Wei‑Lin and Huang, Chao‑Wei and Meng, Yu and Chen, Yun‑Nung},
  booktitle    = {Findings of the Association for Computational Linguistics: EMNLP 2024},
  pages        = {16612--16631},
  year         = {2024}
}

@article{von2025vox,
  title        = {{Vox Populi, Vox AI? Using Large Language Models to Estimate German Vote Choice}},
  author       = {von der Heyde, Leah and Haensch, Anna‑Carolina and Wenz, Alexander},
  journal      = {Social Science Computer Review},
  year         = {2025},
  publisher    = {SAGE Publications Sage CA: Los Angeles, CA}
}

@article{wang2025large,
  title        = {{Large Language Models That Replace Human Participants Can Harmfully Misportray and Flatten Identity Groups}},
  author       = {Wang, Angelina and Morgenstern, Jamie and Dickerson, John P.},
  journal      = {Nature Machine Intelligence},
  pages        = {1--12},
  year         = {2025},
  publisher    = {Nature Publishing Group UK London}
}

@article{zhang2025personalization,
  title        = {{Personalization of Large Language Models: A Survey}},
  author       = {Zhehao Zhang and Ryan A. Rossi and Branislav Kveton and Yijia Shao and Diyi Yang and Hamed Zamani and Franck Dernoncourt and Joe Barrow and Tong Yu and Sungchul Kim and Ruiyi Zhang and Jiuxiang Gu and Tyler Derr and Hongjie Chen and Junda Wu and Xiang Chen and Zichao Wang and Subrata Mitra and Nedim Lipka and Nesreen K. Ahmed and Yu Wang},
  journal      = {Transactions on Machine Learning Research},
  issn         = {2835-8856},
  year         = {2025},
  url          = {https://openreview.net/forum?id=tf6A9EYMo6},
  note         = {Survey Certification}
}

% \section{Language Resource References}
% \label{lr:ref}
% \bibliographystylelanguageresource{lrec2026-natbib}
% \bibliographylanguageresource{languageresource}

% Appendices are not permitted during the initial submission phase, as papers should be self-contained and reviewable on their own. However, appendices will be allowed in the final, camera-ready version. Each camera-ready version may include an appendix, up to ten (10) pages long.

 \appendix
 \section{\emph{German General Social Survey Personas}}
\label{app:ggp}

\begin{figure*}[ht]
    \centering
    \includegraphics[width=\linewidth]{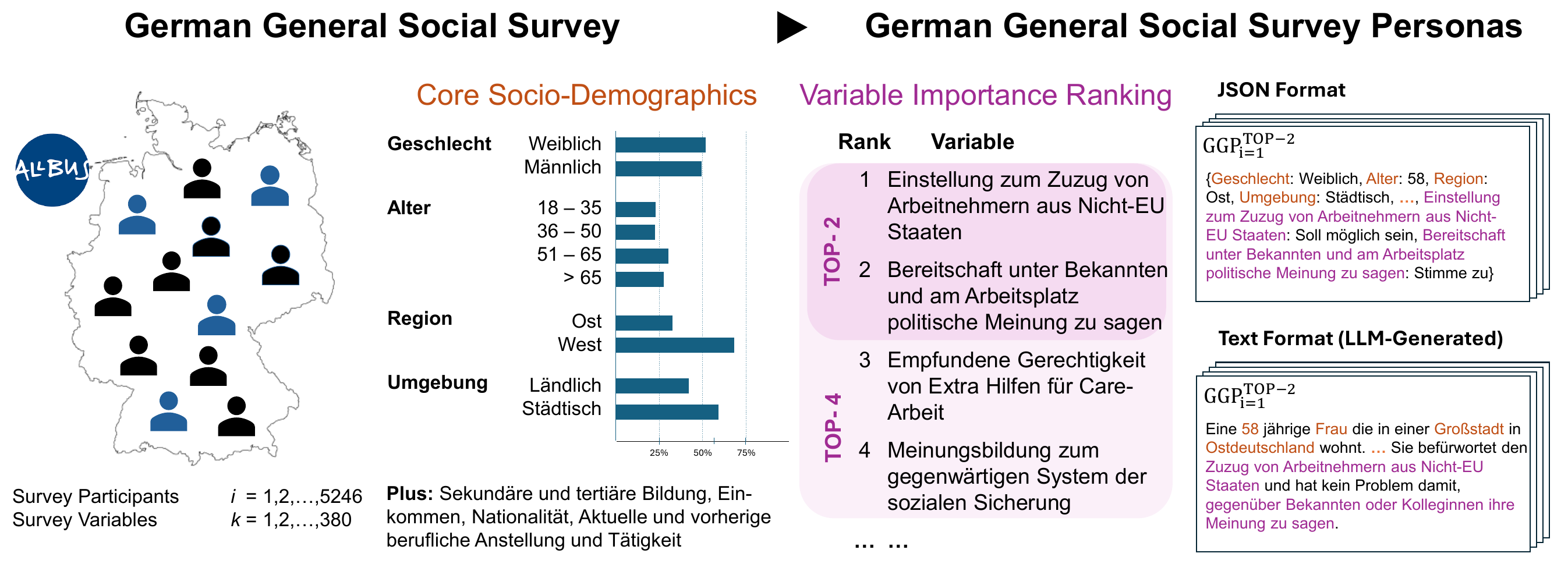}
    \caption{
    \textbf{German Version of Figure 1:} Grounding the German General Social Survey Personas (GGSS Personas) in the ALLBUS survey.}
    \label{fig:figure_1german}
\end{figure*}

Figure~\ref{fig:figure_1german} presents the original survey items and persona descriptions that have been translated to English for Figure~\ref{fig:figure_1}.

\begin{figure}[h]
    \centering
    \includegraphics[width=\columnwidth]{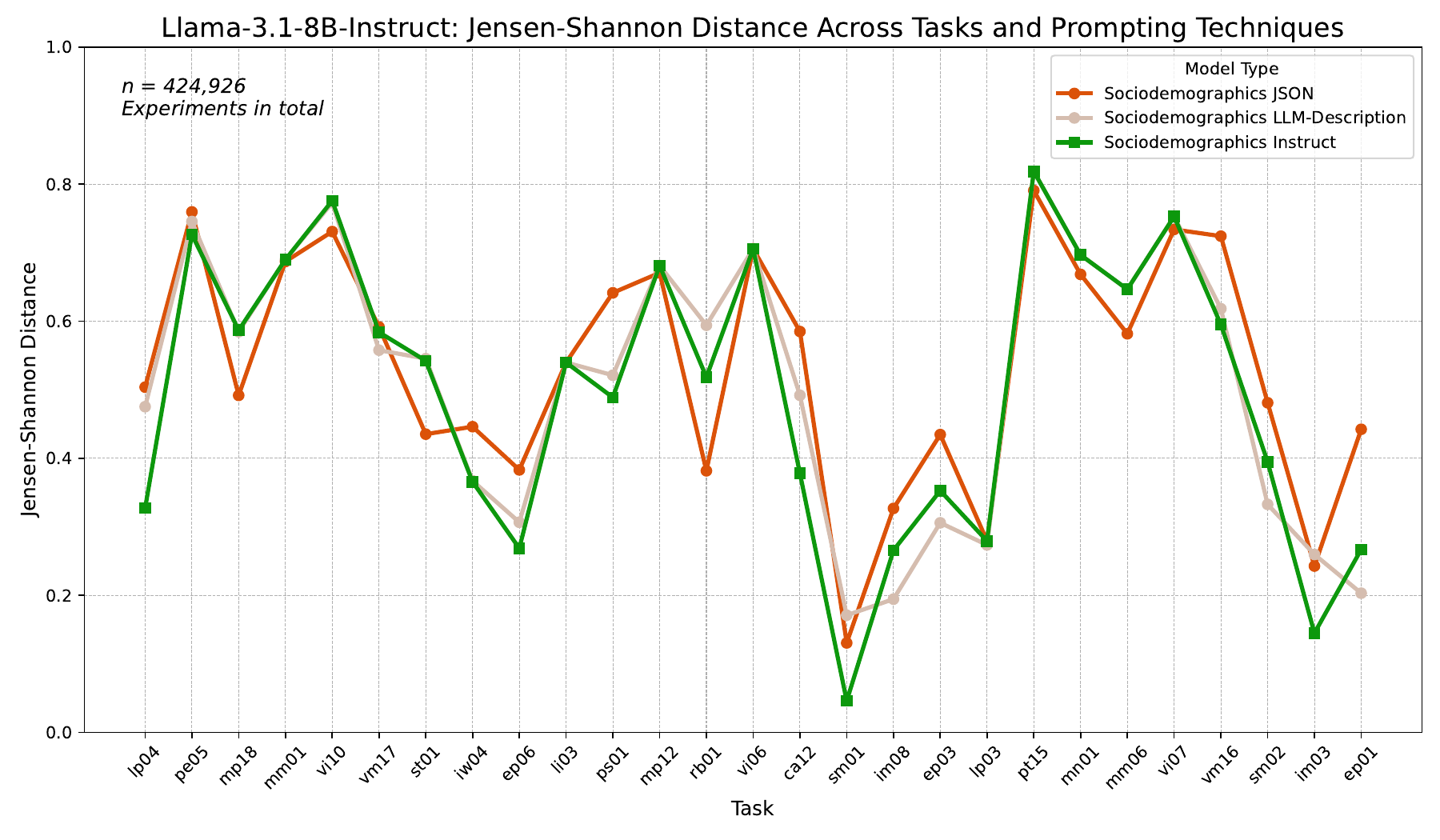}
    \caption{\textbf{Alignment Comparison for different Persona Prompting Techniques.} Except for some tasks, we do not observe consistent and large differences in alignment depending on the exact prompting strategy. In our test, it did make no to very little difference whether the persona attributes are introduced with a a json format, as a description of a person, such as "\textit{Take the view of the following person that ...}". or as a direct instruction, such as "\textit{You are a 50 year old male from western Germany}".}
    \label{fig:jsd_persona_prompt_difference}
\end{figure}

\paragraph{Persona Generation}
We do not observe large differences in response distribution alignment between using different persona formats. Since there is no format that consistently outperforms the others, we decided to conduct our experiments with the personas in \textbf{JSON}-format. Figure \ref{fig:jsd_persona_prompt_difference} underlines this decision exemplarily for \texttt{Llama-3.1-8B-Instruct}.

\section{Preliminary Analysis}

\subsection{Experimental Setup}
\label{app:ex_setup}

\paragraph{Model Selection}

Table~\ref{tab:model_releases} shows the dates of knowledge-cutoff and release for the LLMs used in our preliminary analysis.

\begin{table}[h]
    \centering
    \footnotesize
    \begin{tabular}{lcc}
        \hline
        \textbf{Model} & \textbf{K-Cutoff} & \textbf{R-Date} \\
        \hline
        \textsc{Llama-3.1-8B Instruct} & 12/2023 & 23/07/2024 \\
        \textsc{Llama-3.3-70B-Instruct} & 12/2023 & 06/12/2024 \\
        \textsc{Mistral-7B-Instruct-v0.3} & unknown & 22/05/2024 \\
        \textsc{Qwen3-8B} & unknown & 27/04/2025 \\
        \textsc{gemma-3-12B-it} & 08/2024 & 10/03/2025 \\
        \hline
    \end{tabular}
    \caption{
    \textbf{LLM release dates for empirical evaluation.} All models with a known knowledge cutoff (\textit{K-Cutoff}) were released after the ALLBUScompact date, ensuring no access to training data. For others, the small gap between ALLBUS and model release (\textit{R-Date}) makes data access highly unlikely.
    }
    \label{tab:model_releases}
\end{table}

\paragraph{Response Generation}

Below, we provide an example of a German persona prompt, featuring the core set of socio-demographic attributes as well as the two attributes identified to be most important (TOP-2) according to our variable importance ranking.

\begin{quote}
    "Nehme die Perspektive der folgenden Person ein: {'Wie alt sind Sie?': '50', 'Wohnen Sie in Ost- oder Westdeutschland?': 'Westdeutschland', 'Was ist Ihr höchster Schulabschluss?': '(3) Mittlere Reife', 'In welcher Wohnumgebung wohnen Sie?': 'Großstadt', 'Wie hoch ist Ihr monatliches Nettoeinkommen?': '(11) 1250 - 1374 Euro', 'Wie beschreiben Sie Ihre aktuelle berufliche Tätigkeit?': 'Montageberufe', 'Wie beschreiben Sie Ihre letzte berufliche Tätigkeit?': 'Nicht zutreffend', 'Welches Geschlecht haben Sie?': 'Männlich', 'Welchen Universitätsabschluss besitzen Sie?': 'Kein Universitätsabschluss', 'Wie ist Ihre Einstellung zum Zuzug von Arbeitnehmern aus Nicht-EU-Staaten?': '(2) Zuzug Begrenzen', 'Käme es für Sie in Frage, Ihre politische Meinung im Bekanntenkreis und am Arbeitsplatz zu sagen, um Einfluss zu nehmen?': '(1) Ja'} Welche der Antwortmöglichkeiten ist die Reaktion der Person auf folgende Frage: Meinung zur Verantwortung, Kinder zu bekommen Antwortmöglichkeiten: ['1: BIN DERS.MEINUNG', '2: BIN ANDERER MEINUNG'] Antwort: ("
\end{quote}

This prompt structure instructs the LLM to take on the perspective of the persona described next, provides the survey question the model is supposed to answer as well as the available response options, and sets the model to up to generate its response in the required format.

In this example, the persona is a 50-year-old male from West Germany with an intermediate school leaving certificate, living in a large city and working in assembly with a net monthly income of 1250 - 1374 Euros, supporting limiting immigration from non-EU countries, and willing to voice his political opinions among friends and at work. The prompt then asks for this persona's reaction to a question about the responsibility of having children (variable name: lp04), with the options being "agree" or "disagree".

The English translation of the provided prompt is the following: 
\begin{quote}
    "Assume the perspective of the following person: {'How old are you?': '50', 'Do you live in East or West Germany?': 'West Germany', 'What is your highest school leaving certificate?': '(3) Intermediate school leaving certificate', 'In which residential environment do you live?': 'Large city', 'What is your monthly net income?': '(11) 1250 - 1374 Euro', 'How do you describe your current occupation?': 'Assembly jobs', 'How do you describe your last occupation?': 'Not applicable', 'What is your gender?': 'Male', 'What university degree do you have?': 'No university degree', 'What is your attitude towards the immigration of workers from non-EU countries?': '(2) Limit immigration', 'Would you consider expressing your political opinion to friends and at work to exert influence?': '(1) Yes'} Which of the answer options is the person's reaction to the following question: Opinion on the responsibility to have children Answer options: ['1: I AGREE', '2: I DISAGREE'] Answer: (".
\end{quote}

\paragraph{Decoding Constraints and Invalid Responses}
We generated responses from the whole token distribution. We did not apply any decoding constraints such as the Top-P, Top-K sampling or temperature setting. We do acknowledge that these might impact the response generation. 

A response is valid when the language model generates only one token that reflects one response option on the answer option scale. Table~\ref{tab:missings_topk_personas_percentage} shows the rates with which the different LLMs used for generating synthetic response distributions failed to generate valid responses.

\begin{table}[h]
    \centering
    \footnotesize
    \begin{tabular}{lc}
        \hline
        \textbf{Model} & \textbf{Invalid (\%)} \\
        \hline
        \textsc{Llama-3.1-8B-Instruct} & 1.44 \\
        \textsc{Llama-3.3-70B-Instruct} & 8.75 \\
        \textsc{Mistral-7B-Instruct-v0.3} & 3.73 \\
        \textsc{Qwen3-8B} & 9.91 \\
        \textsc{gemma-3-12b-it} & 1.98 \\
        \hline
    \end{tabular}
    \caption{\textbf{LLMs Invalid Responses by Model for TOP-$K$ Personas Only.} \texttt{Llama-3.1-8B-Instruct} and \texttt{gemma-3-12b-it} have the lowest rates of answering with invalid responses (e.g. refusal, wrong response label, no answer). \texttt{Qwen3-8B} is the most unreliable respondent. Invalid responses increase as arbitrary or oversampled populations are used instead of the representative population.}
    \label{tab:missings_topk_personas_percentage}
\end{table}

\paragraph{Infrastructure}
We carried out the experiments of predicting attributes with GGSS Personas's synthetic responses on a high-performance computing cluster and a local server equipped with NVIDIA H100 (80GB) GPUs and Blackwell GPUs. 
%The total runtime on the computing cluster for **********. 
To accommodate larger models on available hardware, we applied 8-bit quantization to \texttt{Llama-3.3-70B-Instruct}. Smaller models were run without quantization.

Further, we use \textit{vLLM} for more efficient memory management and higher throughput as this package enables sufficient batching of many requests at a time \cite{kwon2023efficient}.

\section{Evaluation}
\label{app:evaluation}
% \subsection{Baseline Comparison}

% \begin{figure*}[h!]
%     \centering
%     \includegraphics[width=\linewidth]{figures/Figure5_baselines_gemma-3-12b-it.pdf}
%     \caption{\textbf{Alignment comparison to baseline persona collections.} This table compares the average Jensen-Shannon Distance across all three tasks in each topic to the ground truth distribution from ALLBUS.}
%     \label{fig:Figure5_gemma_12B}
% \end{figure*}

% \begin{figure*}[h!]
%     \centering
%     \includegraphics[width=\linewidth]{figures/Figure5_baselines_Llama-3.1-8B-Instruct.pdf}
%     \caption{\textbf{Alignment comparison to baseline persona collections.} This table compares the average Jensen-Shannon Distance across all three tasks in each topic to the ground truth distribution from ALLBUS.}
%     \label{fig:Figure5_llama_8B}
% \end{figure*}

% \begin{figure*}
%     \centering
%     \includegraphics[width=\linewidth]{figures/Figure5_baselines_Mistral-7B-Instruct-v0.3.pdf}
%     \caption{\textbf{Alignment comparison to baseline persona collections.} This table compares the average Jensen-Shannon Distance across all three tasks in each topic to the ground truth distribution from ALLBUS.}
%     \label{fig:Figure5_mistral_7B}
% \end{figure*}

% \begin{figure*}
%     \centering
%     \includegraphics[width=\linewidth]{figures/Figure5_baselines_Qwen3-8B.pdf}
%     \caption{\textbf{Alignment comparison to baseline persona collections.} This table compares the average Jensen-Shannon Distance across all three tasks in each topic to the ground truth distribution from ALLBUS.}
%     \label{fig:Figure5_qwen_8B}
% \end{figure*}

\subsection{Validation of Full-Text Persona Generation}
\label{app:human_validation}
We generate full-text personas based on the JSON-format persona descriptions with \texttt{Gemini-2.5-flash}. Although new, state-of-the-art LLMs are powerful and capable of dealing with large context windows, the generated full-text personas must be validated. LLMs are prone to hallucinations or misinterpreting the meaning of a response to a survey item when generating a full-text persona description. Although the maximum input token size of \texttt{Gemini-2.5-flash} exceeds 1 million tokens \footnote{see \href{https://shorturl.at/8UNN2}{Gemini-2.5-flash Documentation}}, more content in the context window might increase the likelihood of errors, hallucinations, or misinterpretation.

% Therefore, two coauthors of this paper manually and independently inspected a sample of 80 personas (20 randomly-sampled personas per TOP-2, -4, -8 and -16 collection) for hallucinations, misrepresentation or omission of individual attributes. We find that 69 of 80 (or 86.25\%) full-text personas are accurate depictions of the JSON-personas, with errors occurring rarely through misrepresentation, e.g., of double negations or complex temporal orders, or through the omission of single attributes.

Apart from the minor misrepresentations and omissions found in the manually validated persona descriptions with up to 16 included attributes (TOP-2,-4,-8 and -16), we anecdotally observed that full-text personas with the largest sets of attributes (TOP-380) show major flaws. For multiple generated persona descriptions we observe generation loops, i.e., situations in which individual sentences in the persona description are repeated until the maximum token limit is reached. Therefore, we recommend using full-text personas with fewer attributes to avoid the use of these types of faulty personas.

\subsection{Hyperparameters for Random Forest Classifier}
\label{app:hyperparameters}

The Random Forest models utilized for classification tasks throughout all experiments were instantiated from the \texttt{scikit-learn} library's \texttt{RandomForestClassifier} class. We selected the hyperparameters according to the findings of \citet{miranda2025simulating} who used the classifier in a similar task. To ensure consistency and comparability across experiments, the following hyperparameters were fixed:

\begin{itemize}
\item \texttt{n\_estimators = 100}: The number of trees in the forest.
\item \texttt{max\_depth = 10}: The maximum depth of each tree.
\item \texttt{min\_samples\_split = 2}: The minimum number of samples required to split an internal node.
\item \texttt{min\_samples\_leaf = 2}: The minimum number of samples required to be at a leaf node.
\item \texttt{random\_state}: A specific integer value was used for the \texttt{random\_state} parameter in each experimental run to ensure reproducibility of the results.
\end{itemize}

These parameters were selected based on preliminary experimentation to provide a reasonable balance between model complexity and generalization. During these initial tests, increasing the number of estimators beyond 100 yielded no measurable gains in performance. Conversely, increasing \texttt{max\_depth} indicated potential overfitting. The goal of this configuration was to define a stable, conservative baseline classifier for comparison with the LLMs, avoiding overfitting the Random Forest for each specific topic.

% \section{Glossary}
% \label{glossary}

% \begin{itemize}
%     \item \textbf{Attribute}: An attribute is one survey variable that becomes part of a persona description. A persona can be constructed of multiple attributes. The more attributes a persona has, the more detailed and nuanced the persona description will be.
%     \item \textbf{Persona}: A persona is a abstraction of a real-life person and contains various attributes. A persona can be generated in different formats, e.g. json-structured, template-format or a full-text persona description.
% \end{itemize}

\section{Survey Data}
\label{survey_data}
The following table lists the outcome variables used to evaluate the \emph{GGSS Personas}. These variables were left out when generating the persona descriptions to avoid introducing the relevant information of the individual response directly into the persona prompt. 
\onecolumn
\begin{longtable}{m{1cm}m{.8cm}m{3.6cm}m{3.6cm}m{2.5cm}m{2.5cm}}
    \footnotesize

    \textbf{Cat.} & \textbf{Var.} & \textbf{German Question} & \textbf{English Question} & \textbf{German Response} & \textbf{English Response} \\
    \hline
    \endfirsthead

    \\
    econ. &
    ep01 &
    Wie beurteilen Sie ganz allgemein die heutige wirtschaftliche Lage in Deutschland? &
    How do you generally assess the current economic situation in Germany? &
    [Sehr gut; Gut; Teils gut / teils schlecht; Schlecht; Sehr schlecht] &
    [Very good; Good; Partly good / partly bad; Bad; Very bad] \\
    \hline
    econ. &
    ep03 &
    Wie beurteilen Sie Ihre eigene wirtschaftliche Lage heute? &
    How do you assess your own economic situation today? &
    [Sehr gut; Gut; Teils gut / teils schlecht; Schlecht; Sehr schlecht] &
    [Very good; Good; Partly good / partly bad; Bad; Very bad] \\
    \hline
    econ. &
    ep06 &
    Was glauben Sie, wie wird Ihre eigene wirtschaftliche Lage in einem Jahr sein? &
    What do you think your own economic situation will be like in a year? &
    [Wesentlich besser als heute; Etwas besser als heute; Gleichbleibend; Etwas schlechter als heute; Wesentlich schlechter als heute] &
    [Significantly better than today; Somewhat better than today; About the same; Somewhat worse than today; Significantly worse than today] \\
    \hline
    ethno. &
    mp12 &
    Auf einer Skala von 1 (stimme überhaupt nicht zu) bis 7 (stimme voll und ganz zu), inwieweit stimmen Sie folgender Aussage zu: "Die Ausländer in Deutschland tragen dazu bei, den Fachkräftemangel zu beheben." &
    On a scale from 1 (strongly disagree) to 7 (strongly agree), to what extent do you agree with the following statement: “Foreigners in Germany help to alleviate the shortage of skilled workers.” &
    7-Punkte Skala (1 := Stimme überhaupt nicht zu --- 7 := Stimme voll und ganz zu) &
    7-point scale (1 = Strongly disagree --- 7 = Strongly agree) \\
    \hline
    ethno. &
    mn01 &
    Auf einer Skala von 1 (überhaupt nicht wichtig) bis 7 (sehr wichtig), wie wichtig sollte Ihrer Meinung nach folgender Umstand bei der Entscheidung über die Vergabe der deutschen Staatsbürgerschaft sein: "Ob die Person in Deutschland geboren ist." &
    On a scale from 1 (not important at all) to 7 (very important), how important do you think the following factor should be in the decision on granting German citizenship: “Whether the person was born in Germany.” &
    7-Punkte Skala (1 := Überhaupt nicht wichtig --- 7 := Sehr wichtig) &
    7-point scale (1 = Not important at all --- 7 = Very important) \\
    \hline
    ethno. &
    mp18 &
    Wegen der in den letzten Jahren nach Deutschland gekommenen Flüchtlinge - denken Sie, es ergeben sich mehr Chancen, mehr Risiken, oder weder noch in Bezug auf das Zusammenleben in unserer Gesellschaft? &
    Considering the refugees who have come to Germany in recent years – do you think this has led to more opportunities, more risks, or neither with regard to living together in our society? &
    [Deutlich mehr Risiken; Eher mehr Riskien; Weder noch; Eher mehr Chancen; Deutlich mehr Chancen] &
    [Significantly more risks; Somewhat more risks; Neither; Somewhat more opportunities; Significantly more opportunities] \\
    \hline
    lifesty. &
    li03 &
    Auf einer Skala von 1 (unwichtig) bis 7 (sehr wichtig), wie wichtig ist Ihnen der Lebensbereich "Freizeit und Erholung"? &
    On a scale from 1 (unimportant) to 7 (very important), how important is the area of life “Leisure and recreation” to you? &
    7-Punkte Skala (1 := Unwichtig --- 7 := Sehr wichtig) &
    7-point scale (1 = Unimportant --- 7 = Very important) \\
    \hline
    lifesty. &
    lp03 &
    Sind Sie derselben oder anderer Meinung mit der Aussage "Egal was manche Leute sagen: Die Situation der einfachen Leute wird nicht besser, sondern schlechter"? &
    Do you agree or disagree with the statement: “No matter what some people say, the situation of ordinary people is not getting better but worse”? &
    [Bin derselben Meinung; Bin anderer Meingung] &
    [Agree; Disagree] \\
    \hline
    lifesty. &
    lp04 &
    Sind Sie derselben oder anderer Meinung mit der Aussage "So wie die Zukunft aussieht, kann man es kaum noch verantworten, Kinder auf die Welt zu bringen"? &
    Do you agree or disagree with the statement: “Given how the future looks, it is hardly justifiable to bring children into the world”? &
    [Bin derselben Meinung; Bin anderer Meingung] &
    [Agree; Disagree] \\
    \hline
    moral. &
    ca12 &
    Was halten Sie von der folgenden Verhaltensweise: "Jemand raucht mehrmals in der Woche Haschisch"? &
    What is your opinion of the following behavior: “Someone smokes hashish several times a week”? &
    4-Punkte Skala (1 := Sehr schlimm --- 4 = Überhaupt nicht shlimm) &
    4-point scale (1 = Very bad --- 4 = Not bad at all) \\
    \hline
    moral &
    vm16 &
    Für Paare, die sich ein Kind wüschen, aber auf natürlichem Wege keines bekommen können - wie beruteilen Sie die folgende Alternative: "Ein Paar verwendet eigene Ei- oder Samenzellen, um mit medizinischer Hilfe ein Kind zu bekommen." &
    For couples who wish to have a child but cannot conceive naturally – how do you evaluate the following alternative: “A couple uses their own egg or sperm cells to have a child with medical assistance.” &
    7-Punkte Skala (-3 := Sehr falsch --- 3 := Sehr richtig) &
    7-point scale (-3 = Very wrong --- 3 = Very right) \\
    \hline
    moral &
    vm17 &
    Für Paare, die sich ein Kind wüschen, aber auf natürlichem Wege keines bekommen können - wie beruteilen Sie die folgende Alternative: "Ein Paar verwendet anonym gespendete Ei- oder Samenzellen, um mit medizinischer Hilfe ein Kind zu bekommen." &
    For couples who wish to have a child but cannot conceive naturally – how do you evaluate the following alternative: “A couple uses anonymously donated egg or sperm cells to have a child with medical assistance.” &
    7-Punkte Skala (-3 := Sehr falsch --- 3 := Sehr richtig) &
    7-point scale (-3 = Very wrong --- 3 = Very right) \\
    \hline
    pol.tend. &
    pt15 &
    Auf einer Skala von 1 (überhaupt kein Vertrauen) bis 7 (sehr großes Vertrauen), wie groß ist das Vertrauen, das Sie "politischen Parteien" entgegenbringen? &
    On a scale from 1 (no trust at all) to 7 (a great deal of trust), how much trust do you have in “political parties”? &
    7-Punkte Skala (1 := Überhaupt kein Vertrauen --- 7 := Sehr großes Vertrauen) &
    7-point scale (1 = No trust at all --- 7 = A great deal of trust) \\
    \hline
    pol.tend. &
    pe05 &
    Inwieweit stimmen Sie folgender Meinung zu: "Die Politiker bemühen sich im Allgemeinen darum, die Interessen der Bevölkerung zu vertreten." &
    To what extent do you agree with the following statement: “Politicians generally try to represent the interests of the population.” &
    4-Punkte Skala (1 := Stimme voll und ganz zu --- 4 := Stimme überhaupt nicht zu) &
    4-point scale (1 = Strongly agree --- 4 = Strongly disagree) \\
    \hline
    pol.tend. &
    ps01 &
    Auf einer Skala von 1 (sehr zufrieden) bis 6 (sehr unzufrieden), wie zufrieden sind Sie - insgesamt betrachtet - mit den gegenwärtigen Leistungen der Bundesregierung? &
    On a scale from 1 (very satisfied) to 6 (very dissatisfied), overall, how satisfied are you with the current performance of the federal government? &
    6-Punkte Skala (1 := Sehr zufrieden --- 6 := Sehr unzufrieden) &
    6-point scale (1 = Very satisfied --- 6 = Very dissatisfied) \\
    \hline
    relig. &
    rb01 &
    Inwieweit stimmen Sie folgender Aussage zu: "Es gibt einen Gott, der sich mit jedem Menschen persönlich befasst." &
    To what extent do you agree with the following statement: “There is a God who concerns Himself personally with every human being.” &
    5-Punkte Skala (1 := Stimme voll und ganz zu --- 5 := Stimme überhaupt nicht zu) &
    5-point scale (1 = Strongly agree --- 5 = Strongly disagree) \\
    \hline
    relig. &
    mm01 &
    Auf einer Skala von 1 (stimme überhaupt nichts zu) bis 7 (stimme voll und ganz zu), inwieweit stimmen Sie folgender Aussage zu: "Die Ausübung des islamischen Glaubens in Deutschland sollte eingeschränkt werden." &
    On a scale from 1 (strongly disagree) to 7 (strongly agree), to what extent do you agree with the following statement: “The practice of the Islamic faith in Germany should be restricted.” &
    7-Punkte Skala (1 := Stimme überhaupt nicht zu --- 7 := Stimme voll und ganz zu) &
    7-point scale (1 = Strongly disagree --- 7 = Strongly agree) \\
    \hline
    relig. &
    mm06 &
    Auf einer Skala von 1 (stimme überhaupt nichts zu) bis 7 (stimme voll und ganz zu), inwieweit stimmen Sie folgender Aussage zu: "Ich habe den Eindruck, dass unter den in Deutschland lebenden Muslimen viele religiöse Fanatiker sind." &
    On a scale from 1 (strongly disagree) to 7 (strongly agree), to what extent do you agree with the following statement: “I have the impression that among Muslims living in Germany there are many religious fanatics.” &
    7-Punkte Skala (1 := Stimme überhaupt nicht zu --- 7 := Stimme voll und ganz zu) &
    7-point scale (1 = Strongly disagree --- 7 = Strongly agree) \\
    \hline
    soc.cap. &
    st01 &
    Manche Leute sagen, dass man den meisten Menschen trauen kann. Andere meinen, dass man nicht vorsichtig genug sein kann im Umgang mit anderen Menschen. Was ist Ihre Meinung dazu? &
    Some people say that most people can be trusted. Others think that you can’t be too careful when dealing with other people. What is your opinion on this? &
    [Den meisten Menschen kann man trauen; Man kann nicht vorsichtig genug sein; Das kommt drauf an] &
    [Most people can be trusted; You can’t be too careful; It depends] \\
    \hline
    soc.cap. &
    sm01 &
    Sind Sie derzeit Mitglied in einer Gewerkschaft? &
    Are you currently a member of a trade union? &
    [Ja; Nein] &
    [Yes; No] \\
    \hline
    soc.cap. &
    sm02 &
    Waren Sie früher einmal Mitglied in einer Gewerkschaft? &
    Have you ever been a member of a trade union in the past? &
    [Ja; Nein] &
    [Yes; No] \\
    \hline
    soc.ineq. &
    im03 &
    Auf einer Skala von 1 (sehr wichtig) bis 4 (unwichtig), wie beurteilen Sie die gegenwärtige Wichtigkeit der folgenden Eigenschaften und Umstände für einen Aufstieg in unserer Gesellschaft: "Bildung, Ausbildung". &
    On a scale from 1 (very important) to 4 (unimportant), how do you rate the current importance of the following characteristic or factor for upward mobility in our society: “Education, training.” &
    4-Punke Skala (1 := Sehr wichtig --- 4 := Unwichtig) &
    4-point scale (1 = Very important --- 4 = Unimportant) \\
    \hline
    soc.ineq. &
    im08 &
    Auf einer Skala von 1 (sehr wichtig) bis 4 (unwichtig), wie beurteilen Sie die gegenwärtige Wichtigkeit der folgenden Eigenschaften und Umstände für einen Aufstieg in unserer Gesellschaft: "Leistung, Fleiß". &
    On a scale from 1 (very important) to 4 (unimportant), how do you rate the current importance of the following characteristic or factor for upward mobility in our society: “Achievement, diligence.” &
    4-Punke Skala (1 := Sehr wichtig --- 4 := Unwichtig) &
    4-point scale (1 = Very important --- 4 = Unimportant) \\
    \hline
    soc.ineq. &
    iw04 &
    Inwieweit stimmen Sie folgender Aussage zu: "Der Staat muss dafür sorgen, dass man auch bei Krankheit, Not, Arbeitslosigkeit und im Alter ein gutes Auskommen hat." &
    To what extent do you agree with the following statement: “The state must ensure that people have a decent standard of living even in cases of illness, hardship, unemployment, and old age.” &
    4-Punkte Skala (1 := Stimme voll zu --- 4 := Stimme überhaupt nicht zu) &
    4-point scale (1 = Completely agree --- 4 = Do not agree at all) \\
    \hline
    values &
    vi06 &
    Auf einer Skala von 1 (unwichtig) bis 7 (außerordentlich wichtig), wie wichtig ist Ihnen "Sozial Benachteiligten und gesellschaftlichen Randgruppen helfen"? &
    On a scale from 1 (unimportant) to 7 (extremely important), how important is “Helping socially disadvantaged people and marginalized groups” to you? &
    7-Punkte Skala (1 := Unwichtig --- 7 := Außerordentlich wichtig) &
    7-point scale (1 = Unimportant --- 7 = Extremely important) \\
    \hline
    values &
    vi07 &
    Auf einer Skala von 1 (unwichtig) bis 7 (außerordentlich wichtig), wie wichtig ist Ihnen "Sich und seine Bedürfnisse gegen andere durchsetzen"? &
    On a scale from 1 (unimportant) to 7 (extremely important), how important is “Asserting yourself and your needs against others” to you? &
    7-Punkte Skala (1 := Unwichtig --- 7 := Außerordentlich wichtig) &
    7-point scale (1 = Unimportant --- 7 = Extremely important) \\
    \hline
    values &
    vi10 &
    Auf einer Skala von 1 (unwichtig) bis 7 (außerordentlich wichtig), wie wichtig ist Ihnen "Sich politisch engagieren"? &
    On a scale from 1 (unimportant) to 7 (extremely important), how important is “Being politically active” to you? &
    7-Punkte Skala (1 := Unwichtig --- 7 := Außerordentlich wichtig) &
    7-point scale (1 = Unimportant --- 7 = Extremely important) \\
    \\
    \caption{\textbf{Overview of the variables selected as prediction tasks.} For each of the nine topical areas (\textit{cat.}) we randomly select three variables (\textit{var.}) as prediction tasks. We show the original ALLBUS questions and the corresponding response options (slightly adjusted for readability), as well as their English translations.}
    \label{tab:task_variables}
    
\end{longtable}
\twocolumn

\end{document}